\documentclass{article} 
\usepackage[utf8]{inputenc}
\usepackage[T1]{fontenc}
\PassOptionsToPackage{table,dvipsnames}{xcolor}
\usepackage{iclr2027_conference,times}

\usepackage{amsmath,amsfonts,bm}

\def\eqref#1{equation~\ref{#1}}

\def\1{\bm{1}}

\def\eps{{\epsilon}}

\DeclareMathAlphabet{\mathsfit}{\encodingdefault}{\sfdefault}{m}{sl}
\SetMathAlphabet{\mathsfit}{bold}{\encodingdefault}{\sfdefault}{bx}{n}

\usepackage{hyperref}
\usepackage{url}
\usepackage{graphicx}
\usepackage{wrapfig}
\usepackage{subcaption}
\usepackage{tabularx}
\usepackage{hyunin}
\usepackage[table]{xcolor}
\usepackage{etoolbox}
\usepackage{algorithm2e}
\usepackage{enumitem}

\usepackage{fontawesome5}
\usepackage{hyperref}
\usepackage{xcolor}

\definecolor{linkblue}{RGB}{0,0,238}

\newcommand{\projectlinks}{
\vspace{-1.6em}
\begin{center}
    \small
    \faGithub\;
    \href{https://github.com/SakanaAI/mass}
    {\textcolor{linkblue}{\texttt{code}}}
    \hspace{1em}
    {\color{gray}\rule[-0.15em]{0.35pt}{1.2em}}
    \hspace{1em}
    \faGlobe\;
    \href{https://pub.sakana.ai/mass/}
    {\textcolor{linkblue}{\texttt{blog}}}
\end{center}
\vspace{-0.8em}
}

\usepackage{fvextra}
\usepackage{newunicodechar}
\newunicodechar{’}{\textquoteright}
\newunicodechar{‘}{\textquoteleft}
\newunicodechar{“}{\textquotedblleft}
\newunicodechar{”}{\textquotedblright}
\newunicodechar{—}{\textemdash}
\newunicodechar{–}{\textendash}
\newunicodechar{≥}{\ensuremath{\geq}}
\fvset{fontsize=\footnotesize,breaklines=true,breakanywhere=true,
       frame=single,framesep=4pt,numbersep=5pt}

\title{Recursive Self-Improvement through Multi-Agent Self-Supervision}

\author{
Hyunin Lee \thanks{This work was done during an internship at SakanaAI. Corresponding author: \texttt{hyunin@berkeley.edu}}\\
UC Berkeley\\
\And
Jinglue Xu \\
Sakana AI\\
\And
Jeffrey Seely\\ 
Sakana AI\\
\And
Donghyun Lee\\
UC Berkeley\\
\AND
Somayeh Sojoudi\\
UC Berkeley \\
\And
Matei Zaharia\\
UC Berkeley\\
\And 
Yujin Tang\\
Sakana AI\\
}

\iclrfinalcopy 
\begin{document}

\maketitle

\begin{abstract}
Recursive self-improvement (RSI) of a model on non-verifiable tasks, such as open-ended research, faces a supervision bottleneck when its outputs exceed what even human experts can reliably assess, leaving the model itself (\emph{optimizee}) as the best available optimizer and evaluator.
However, a single model instance struggles to critique and improve its own complex reasoning under this homogeneous loop.
To address this, we propose Multi-Agent Self-Supervision (MASS), an RSI method that alternates between evolutionary workflow optimization and supervised fine-tuning on self-generated trajectories.
Guided by early findings that multi-agent topologies excel at complex reasoning, MASS prompts a single base model to iteratively propose, execute, and self-evaluate multi-agent workflows.
Through an evolutionary search constrained by structural guardrails, the model optimizes these computational-graph-like orchestrations, discovering the most effective distinct roles and information routing for a given task.
Over two MASS cycles with Qwen3.6-27B, the model achieves 1.2–1.6× higher performance per \emph{output tokens} on four open-ended public benchmarks. Because the improved model subsequently acts as a better optimizer and evaluator, this alternating framework enables a continuous, recursive bootstrapping of the model's capabilities. 
Moreover, multi-agent traces are also more training-efficient: a student trained on them outperforms a single-agent student trained on 1.4× more \emph{training tokens}. These findings suggest that jointly learning orchestration and bounded subagent execution from multi-agent trajectories can provide an effective signal for RSI.
\end{abstract}
\projectlinks

\begin{figure}[h]
    \centering
    \includegraphics[width=0.9\linewidth]{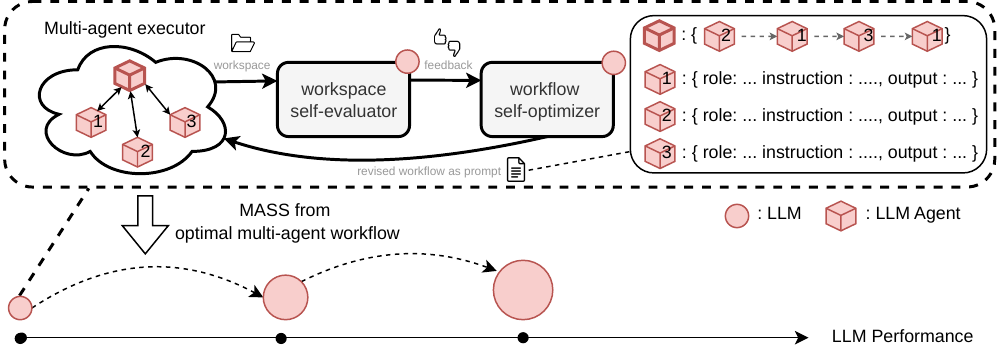}
    \caption{Recursive self-improvement with multi-agent self-supervision (MASS). The agent, evaluator, and optimizer share the same language model (LM). MASS encodes multi-agent coordination as textual information within a \emph{workflow prompt}. In the inner loop, the agent generates a workspace under the current workflow prompt, the self-evaluator provides feedback, and the self-optimizer refines the workflow prompt. The outer loop fine-tunes the shared LM on rollouts from the optimized workflow, then restarts the inner loop with the updated model.
    }
    \label{fig:mainfigure}
\end{figure}
\section{Introduction}

  
Frontier AI systems now contribute to original scientific results, including improved sphere-packing bounds, new gluon-scattering formulas, and experimentally validated nanobody designs \citep{openai2026mathematics,guevara2026single,swanson2025virtual}.
To sustain the scale and pace of this progress, a natural direction is recursive self-improvement (RSI), a framework fundamentally driven by two core components, evaluation and improvement.
However, automating both components at the frontier presents distinct bottlenecks.
On the evaluation side, obtaining reliable supervision is difficult when the work exceeds the expertise of available human evaluators \citep{burns2023weak}, or when feedback arrives too slowly.
For example, in nuclear fusion research, a single tokamak plasma study may require more than 120 million CPU-hours~\citep{maheswaran2026squeeze}.
On the improvement side, advancing the model's capabilities requires training it on high-quality execution data, which in turn depends on discovering better problem-solving workflows to generate that data.
Manually revising task decompositions, tool use, and verification steps to optimize these workflows requires prohibitive effort across tasks and domains \citep{hu2025automated,zhang2025aflow}.
We study this dual challenge of RSI on open-ended research tasks, in a setting where the strongest available model serves interchangeably as the task solver, workflow optimizer, and evaluator.
Although this model-as-evaluator method alleviates the problem on the evaluation side, the central challenge is obtaining useful self-supervision when the model being improved is also its best available supervisor.


This challenge arises because generation, evaluation, and revision share the same learned prior, meaning that self-feedback can reinforce rather than correct errors \citep{huang2024selfcorrect}.
To break this homogeneous constraint, we seek better supervision by reorganizing the model’s working style.
We draw on Minsky’s \emph{Society of Mind}, which views intelligence as emerging from organized interactions among simpler components \citep{minsky1986society}.
This perspective motivates coordinating copies of the model to elicit capabilities that an isolated instance may not reliably exhibit.
For example, when automating quantitative finance research, one model instance might be prompted to do data preprocessing, another to generate trading signals, and a third to run backtests.
The workflow acts as a structured team rather than a single, monolithic generator.
By orchestrating these interactions, we can break down a complex task into simpler pieces, overcome individual blind spots, and then distill their collective experience into shared weights.

Two hypotheses guide us in designing a solution that may elicit these gains and learn efficiently. 
\begin{itemize}
    \item \textbf{Generating higher-quality data.} Optimizing inter-agent information flow can elicit collective capabilities that provide effective self-supervision from a perspective other than incorporating stronger reasoning chains. Building on evidence that optimizing communication structure can improve collective performance \citep{zhuge2024gptswarm,lee2026recursive}, we focus on optimizing what agents communicate, to whom, and when—not simply how many agents participate. 
    \item \textbf{Maximize learning efficiency during fine-tuning.} Jointly training on subagent execution and orchestration enjoys greater learning efficiency than training on a single-agent's long-horizon trajectories. This hypothesis draws inspiration from hierarchical imitation learning, where task decomposition can improve learning efficiency \citep{le2018hierarchical}. In our setting, subagent trajectories demonstrate bounded local execution, while orchestrator trajectories demonstrate delegation, verification, and integration. We conjecture that learning to solve long-horizon tasks from these complementary behaviors at their respective levels of abstraction is easier than learning from one extended trajectory that interleaves both levels.
\end{itemize}

To this end, we propose Multi-Agent Self-Supervision (MASS) for RSI, a method that alternates between workflow optimization and model post-training (Figure~\ref{fig:mainfigure}).
Motivated by findings that multi-agent workflows outperform single-agent execution on open-ended research tasks~\citep{lee2026recursive}, MASS's inner loop refines workflows through an iterative feedback process where the agent executes a multi-agent workflow, the self-evaluator assesses the resulting artifacts, and the optimizer revises the workflow.
The outer loop generates trajectories under the optimized workflows, then jointly fine-tunes the shared model on orchestrator and subagent conversations.
The updated model resumes all three roles in the next cycle.
We evaluate MASS in a controlled setting where the agent, optimizer, and evaluator share the same base model.
Final performance is assessed separately using external strong model judgments withheld from the agents and workflow optimizer (Section~\ref{sec:experiment_setup}).

\begin{figure}[h]
    \centering
    \includegraphics[width=1.0\linewidth]{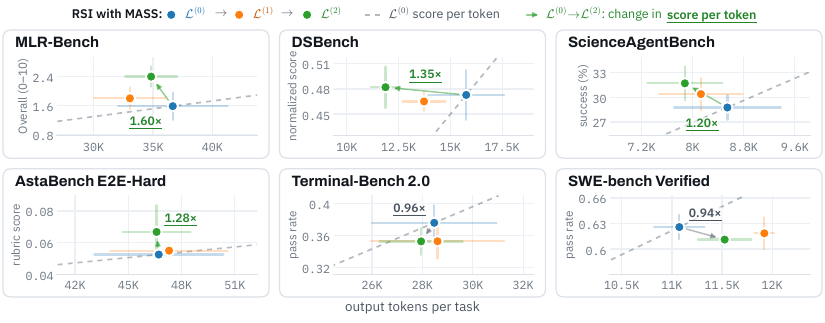} 
    \caption{Score versus output tokens per task on six public benchmarks. $\mathcal{L}^{(0)}$ is the base LM
(Qwen3.6-27B); $\mathcal{L}^{(1)}$ and $\mathcal{L}^{(2)}$ follow one and two cycles of RSI with MASS, all run in
the same agent scaffold (qwen-code). Arrows go from $\mathcal{L}^{(0)}$ to $\mathcal{L}^{(2)}$, labelled with the
ratio of their score per output token: on the four research benchmarks (top row, bottom left), $\mathcal{L}^{(2)}$
gains 1.2--1.6$\times$. Bars show standard deviations over three trials.}
    \label{fig:teaser}
\end{figure}

We train Qwen3.6-27B as the base model on multi-agent task-solving trajectories from eight synthetic research tasks spanning finance, robotics, and pharmacy, and evaluate MASS on three test synthetic research tasks, and six public benchmarks: MLR-Bench~\citep{chen2026mlr}, ScienceAgentBench~\citep{chen2025scienceagentbench}, DSBench~\citep{jing2025dsbench}, AstaBench~\citep{bragg2026astabench}, Terminal-Bench~2.0~\citep{merrill2026terminal}, and SWE-bench Verified~\citep{jimenez2024swe}.  Across two MASS cycles, the trained model's win rate against the base model on the three test tasks increases from $53.9\%$ to $69.9\%$ (Section~\ref{sec:main_results}, Figure~\ref{tab:agent}), and on the four research benchmarks (MLR-Bench, ScienceAgentBench, DSBench, and AstaBench) its score per output token is $1.2$--$1.6\times$ that of the base model (Figure~\ref{fig:teaser}).  Notably, post-training only on task-solving trajectories improves all three roles in the RSI loop: task solving, workflow optimization, and evaluation (Section~\ref{sec:main_results}, Figure~\ref{fig:combined}). The trained model discovers high-quality workflows faster, and the evaluator accuracy increases from $73\%$ to $93\%$.

We further find that optimized workflows develop more task-specific information handoffs and routing, consistent with our information-flow hypothesis (Section~\ref{sec:hypothesis_information_flow}). The first-cycle multi-agent student achieves a $68.3\%$ win rate against the base model, compared with $64.0\%$ for a single-agent student trained on approximately $1.4\times$ more processed supervised tokens (Section~\ref{sec:teacher_ablation}, Figure~\ref{fig:teacher_token_efficiency}). This suggests that jointly learning orchestration and bounded subagent execution can provide more effective supervision for RSI. Our ablations further identify workflow optimization, rather than evaluation, as a major bottleneck in the homogeneous RSI loop (Section~\ref{sec:optimization_ablations}). Finally, trajectory analysis provides evidence that post-training internalizes information-handoff behaviors (Section~\ref{sec:behavioral_internalization}).

Overall, our work introduces an algorithm that uses multi-agent coordination to generate self-supervision for RSI.




\section{Notations and problem statement}
\label{sec:problem_statement}

\textbf{Task and language model.} Let $\mathcal{T}$ denote a finite set of tasks, with $x_t$ denoting the prompt for task $t\in\mathcal{T}$.
We denote a training subset as $\mathcal{T}_{\mathrm{train}}\subset\mathcal{T}$ and a disjoint test set as $\mathcal{T}_{\mathrm{test}}$.
Let $\cL_\theta$ denote a language model with parameters $\theta$, defining the next-token distribution $p_\theta(z_\ell\mid z_{<\ell})$.
For any textual input $x$, we write $\cL(x)$ for a sample drawn from the model conditioned on $x$.

\textbf{Agent and artifacts.} We write $\cA(\cdot;\cL)$ for a coding agent instantiated with language model $\cL$.
Its harness, including tools and memory, is fixed throughout the paper.
A coding-agent execution on task $t$ is denoted by $(\tau_t,y_t)\sim\cA(x_t;\cL)$, where $\tau_t$ is a single (orchestrator-subagent) trajectory, and $y_t$ is the resulting deliverable workspace, such as python files in the coding space.
For an additional textual input $x'$, we use $\oplus$ to denote prompt concatenation.
Thus, $(\tau'_t,y'_t)\sim\cA(x_t\oplus x';\cL)$ denotes execution with the concatenated prompt $x_t\oplus x'$.

\textbf{Task execution and evaluation.} We use $\cL_{\mathrm{agent}}$, $\cL_{\mathrm{eval}}$, and $\cL_{\mathrm{opt}}$ to denote the language models serving as the agent executor, task evaluator, and workflow optimizer, respectively.
In our homogeneous setting,
$\cL_{\mathrm{agent}}
=\cL_{\mathrm{eval}}
=\cL_{\mathrm{opt}}$.
For two outputs $y_1$ and $y_2$ and an evaluation prompt $x_{\mathrm{eval}}$, we write $y_1 \succ y_2$ if $\cL_{\mathrm{eval}}$ prefers $y_1$ over $y_2$.
We denote the corresponding rationale (or feedback) sampled from $\cL_{\mathrm{eval}}$ as $d \sim \cL_{\mathrm{eval}}(y_1,y_2;x_{\mathrm{eval}})$.

\textbf{Goal definition.} Let $\cL^{(g)}$ denote the language model in MASS after $g=0,\cdots,R$ generations of RSI, our goal is to increase the probability of $y^{(g)} \succ y^{(<g)}$ from the perspective of an external strong language model $\cL_{\mathrm{ext}}$ (for the purpose of fair comparisons), where $(\tau^{(g)}, y^{(g)}) \sim \cA(x_t, \cL^{(g)})$.
\section{Method}

\paragraph{Overview.} 

MASS alternates between multi-agent workflow optimization (Algorithms~\ref{alg:MASS_1}) and model fine-tuning that internalizes the workflows (Algorithms~\ref{alg:MASS_2}).
At generation $g$, the current model $\cL^{(g)}$ serves as the task-solving agent, workflow optimizer, and evaluator.
Algorithm~\ref{alg:MASS_1} first produces task-specific retained workflows using fixed model weights.
Algorithm~\ref{alg:MASS_2} then generates and self-selects trajectories under these workflows and fine-tunes the same model to obtain $\cL^{(g+1)}$.
The updated model assumes all three roles in the next generation.
Thus, MASS forms a recursive loop in which the current model first improves how copies of itself coordinate, then learns from the resulting coordinated trajectories.
The next generation repeats the same process with the updated weights. After $R$ iterations, we evaluate $\cL^{(g)}$ on $\cT_{\text{test}}$ with $\cL_{\mathrm{ext}}$ for all $g$ and report the test score.

{\RestyleAlgo{ruled}

\begin{algorithm}[h]
\caption{Recursive self-improvement with MASS}
\label{alg:MASS}

\KwIn{
training tasks $\mathcal{T}_{\mathrm{train}}$;
test tasks $\mathcal{T}_{\mathrm{test}}$;
coding agent $\mathcal{A}$;
initial trainable language model $\cL^{(0)}$;
number of MASS cycles $R$
}

\For{$g=0,1,\ldots,R-1$}{
    Set
    $\cL_{\mathrm{agent}}
    =\cL_{\mathrm{eval}}
    =\cL_{\mathrm{opt}}
    =\cL^{(g)}$\;

    Obtain optimal workflows
    $\{w_t^{*,(g)}\}_{t\in\mathcal{T}_{\mathrm{train}}}$
    using Algorithm~\ref{alg:MASS_1}\;

    Obtain
    $\cL^{(g+1)}$
    by applying Algorithm~\ref{alg:MASS_2}
    to $\cL^{(g)}$ and
    $\{w_t^{*,(g)}\}_{t\in\mathcal{T}_{\mathrm{train}}}$\;

}
\Return $\cL^{(R)}$\;
\end{algorithm}

}

\textbf{Algorithm~\ref{alg:MASS_1}. Multi-agent workflow optimization.} Defining multi-agent workflow in the prompt is demonstrated to be an effective way for LLMs to follow the organization instruction and act accordingly~\citep{nielsen2026learning, tang2026sakana}.
Following this, we define a workflow as a textual specification of an orchestrator--subagent architecture that is appended to the task prompt and guides how the orchestrator delegates work, exchanges information, and invokes subagents.
Specifically, for task $t$, we represent the workflow as $w_t:=\{\mathbf h_t,{(\mathbf r_{t,j},\mathbf u_{t,j},\mathbf c_{t,j})}_{j=1}^{m_t}\}$, where $m_t$ is the number of subagents.
Here, $\mathbf r_{t,j}$ specifies subagent $j$'s role, $\mathbf u_{t,j}$ its instructions, and $\mathbf c_{t,j}$ its output contract.
The contract specifies what information the subagent returns to the orchestrator, while $\mathbf h_t$ specifies the order of subagent calls.
We concatenate $w_t$ to the task prompt $x_t$, and execute the coding agent as $\cA(x_t\oplus w_t;\cL_{\text{agent}})$.
The orchestrator and all subagents share the same $\cL_\text{agent}$.
A simplified workflow format is shown below; complete examples appear in Appendices~\ref{app:crown51_full} and~\ref{app:crown202_full}.

\begin{tcolorbox}[
    colback=gray!5,
    colframe=black!70,
    title=Multi-agent workflow $w$,
    fonttitle=\small,
    sharp corners
]
\small
\begin{itemize}[nosep]
    \item \texttt{orchestrator}:
    $\mathbf h =
    \{\texttt{subagent}_2
    \rightarrow \texttt{subagent}_{m}
    \rightarrow \cdots
    \rightarrow \texttt{subagent}_1\}$

    \item $\texttt{subagent}_1:
    \{
    \text{role}: \mathbf r_1,\;
    \text{instruction}: \mathbf u_1,\;
    \text{output contract}: \mathbf c_1
    \}$

    \item $\cdots$

    \item $\texttt{subagent}_m:
    \{
    \text{role}: \mathbf r_m,\;
    \text{instruction}: \mathbf u_m,\;
    \text{output contract}: \mathbf c_m
    \}$
\end{itemize}
\end{tcolorbox}

For each training task $t\in\mathcal{T}_{\mathrm{train}}$, workflow
optimization maintains an independent sequence
$\{w_t^{(i)}\}_{i\geq 0}$ and a best so-far workflow $w_t^*$,
as determined by $\cL_{\mathrm{eval}}$.
Executing workflow $w_t^{(i)}$ produces
$(\tau_t^{(i)},y_t^{(i)})
\sim \cA(x_t\oplus w_t^{(i)};\cL_{\mathrm{agent}})$.
At iteration $i$, $\cL_{\mathrm{eval}}$ compares the current output
$y_t^{(i)}$ with the output $y_t^*$ produced by the best so-far workflow
$w_t^*$ and returns feedback
$d_{t,i}=
\cL_{\mathrm{eval}}(y_t^{(i)},y_t^*;x_{\mathrm{eval}})$.
If $y_t^{(i)}\succ y_t^*$, we update
$w_t^*\leftarrow w_t^{(i)}$ and
$y_t^*\leftarrow y_t^{(i)}$. The feedback $d_{t,i}$ is then added to the task-specific history, which $\cL_{\mathrm{opt}}$ uses, together with the optimization prompt $x_{\mathrm{opt}}$, to propose the next workflow $w_t^{(i+1)}$.
This resembles an evolution process with an elite size one \citep{risi2026neuroevolution}.

To prioritize improvements in multi-agent coordination,
$x_{\mathrm{opt}}$ explicitly emphasizes the communication structure
$(\mathbf h_t,\{\mathbf c_{t,j}\}_{j=1}^{m_t})$
over the role and instruction components
$\{(\mathbf r_{t,j},\mathbf u_{t,j})\}_{j=1}^{m_t}$.
Thus, $w_t^*$ denotes the best self-judged workflow under sequential in-loop pairwise evaluation, rather than a certified global optimum.
We show the overview of $x_{\mathrm{opt}}$ below, Algorithm~\ref{alg:MASS_1} in Appendix \ref{app:algorithm1} summarizes the procedure.

\begin{tcolorbox}[
    colback=gray!5,
    colframe=black!70,
    title=Optimization prompt $x_\mathrm{opt}$,
    fonttitle=\small,
    sharp corners
]
\small
\begin{itemize}[nosep]
    \item \texttt{Task}:
    $\mathbf t =
    \text{``Find trading signal with backtest results ...''}$

    \item \texttt{Current best workflow}: $w_t^*$

    \item \texttt{Structure emphasis}: $(\mathbf h_t,\{\mathbf c_{t,j}\}_{j=1}^{m_t}) \text{ and } \{(\mathbf r_{t,j},\mathbf u_{t,j})\}_{j=1}^{m_t}$
    \item \texttt{Evaluator feedback history}: $\{d_{t,1},\ldots,d_{t,i}\}$
\end{itemize}
\end{tcolorbox}

\paragraph{Algorithm~\ref{alg:MASS_2}. Workflow internalization by supervised fine-tuning.}

We next internalize the behavior induced by the workflows $w_t^*$ obtained from Algorithm~\ref{alg:MASS_1} through supervised fine-tuning (SFT).
For each training task $t\in\mathcal{T}_{\mathrm{train}}$, we execute $\cA(x_t\oplus w_t^*;\cL_{\mathrm{agent}})$ independently $M$ times, yielding candidate trajectory--workspace pairs $\{(\tau_{t,j},y_{t,j})\}_{j=1}^{M}$.
Within each task, $\cL_{\mathrm{eval}}$ compares pairs of deliverable workspaces $y_{t,{j_1}}$ and $y_{t,{j_2}}$, where $j_1\neq j_2\in[M]$, in a pairwise tournament.
A Bradley--Terry model converts these pairwise judgments into workspace-level scores.
We rank the trajectory--workspace pairs by score, use the top $K-1$ trajectories $\{\tau_{t,j}\}_{j=1}^{K-1}$ from each task as the SFT training set $\cD_{\mathrm{train}}$, and reserve the $K$th trajectory for the validation set $\cD_{\mathrm{val}}$. We remove $w^*_t$ from the prompt for dataset construction.

Trajectories in $\cD_{\mathrm{train}}$ and $\cD_{\mathrm{val}}$ are partitioned into windows of at most $L_{\max}$ tokens, with preceding context optionally retained but masked from the loss.
Let $\mathcal{D}$ denote the resulting set of SFT windows.
For a window $z=(z_1,\ldots,z_{|z|})\in\mathcal{D}$, let
$\mu_\ell=1$ if $z_\ell$ is a supervised assistant target and
$\mu_\ell=0$ for prompt tokens, tool outputs, and masked context.
Let $q$ denote the sampling distribution over stored windows.
The SFT objective is
$\ell_{\mathrm{SFT}}(\theta)
=
-\mathbb{E}_{z\sim q}
\left[
\frac{
\sum_{\ell=1}^{|z|}
\mu_\ell \log p_\theta(z_\ell\mid z_{<\ell})
}{
\sum_{\ell=1}^{|z|}
\mu_\ell
}
\right]$.
Algorithm~\ref{alg:MASS_2}  in Appendix \ref{app:algorithm2} summarizes the procedure.


\section{Experiments}
\label{sec:experiments}

\subsection{Experimental setup}
\label{sec:experiment_setup}


\noindent\textbf{Models and tasks.}
Our base model $\cL^{(0)}$ is Qwen3.6-27B-FP8, and we use
\texttt{qwen-code}~0.20.0 as the coding agent $\cA$.
For the homogeneous first-cycle setting, we set
$\cL_{\mathrm{agent}}
=\cL_{\mathrm{eval}}
=\cL_{\mathrm{opt}}
=\cL^{(0)}$.
We use 12 synthetic research-programming (long-horizon) tasks (private benchmark) generated by GPT-5.5, with four tasks each in finance, robotics, and pharmacy
\citep{lee2026recursive}. We designate nine tasks as train tasks $\cT_{\text{train}}$ and three as part of test
tasks, $\mathcal{T}_{\mathrm{test}}$,
with one test task per domain.
If Algorithm~\ref{alg:MASS_1} fails to find an optimal multi-agent workflow that
outperforms its reference for a candidate training task, that task is
excluded from teacher-data collection in
Algorithm~\ref{alg:MASS_2}.
In our experiments, this leaves eight tasks for post-training.
Examples of task prompts $x_t$ and optimal workflows $w_t^*$
appear in Appendices~\ref{app:task_examples}--\ref{app:crown_examples}.

\noindent\textbf{Test tasks.}
Test tasks $\mathcal{T}_{\mathrm{test}}$ consist of ScienceAgentBench (102 tasks) \citep{chen2025scienceagentbench}, MLR-Bench (107tasks) \citep{chen2026mlr}, DSBench \citep{jing2025dsbench}, AstaBench E2E-Bench-hard \citep{bragg2026astabench}, Terminal-Bench 2.0 \citep{merrill2026terminal},SWE-bench Verified \citep{jimenez2024swe}, and three synthetic research tasks described above. Those three synthetic tasks are
(1) The finance task asks the agent to backtest an equity long--short
strategy on S\&P~500 data from Yahoo Finance and estimate trading
capacity after accounting for spreads, market impact, and borrowing
costs.
(2) Pharmacy task evaluates inverse-folding models on Protein Data
Bank backbones against random and Markov baselines using sequence
recovery and structure-quality proxies \citep{berman2000protein}.
(3) Robotics task requires constructing a tool-use manipulation
benchmark using RoboTAP \citep{vecerik2024robotap} or custom PyBullet tasks and comparing symbolic
planning with behavior cloning in terms of task success, tool
selection, and failure modes.
Each task requires reproducible code, quantitative results, and a
research report.

\noindent\textbf{Train data and post-training.}
In Algorithm~\ref{alg:MASS_2}, we generate $M=18$ candidate
trajectories per task and select up to $K=16$ teacher trajectories.
One candidate training task is excluded because
Algorithm~\ref{alg:MASS_1} does not find a workflow that outperforms
its reference.
Across the remaining eight training tasks, the final training set
contains $(9-1) \times 15 -1 = 119$ trajectories: up to 15 per task, with one trajectory
excluded because the coding-agent harness terminated prematurely.
More details are documented in
Appendices~\ref{app:training_protocol} and~\ref{app:data_details}.

The 119 teacher trajectories yield 2,435 training windows
(282 orchestrator, 2,054 subagent, and 99 auxiliary), containing
17.2M supervised tokens.
We set $L_{\max}=49{,}152$ tokens, with up to $8{,}192$ tokens of
preceding context retained but masked from the loss.
Because subagent windows substantially outnumber orchestrator windows,
we use a fixed $2{:}1$ subagent-to-orchestrator sampling ratio.   We train the BF16 parent model with a newly initialized LoRA adapter \citep{hu2022lora} using rank 64, scaling factor 128, and dropout
$0.05$. Optimization uses AdamW \citep{loshchilov2019decoupled} with
$\beta_1=0.9$, $\beta_2=0.95$, a peak learning rate of
$3\times10^{-5}$, five warmup steps, and cosine decay to $10\%$ of
the peak learning rate, finally conducted 1,624 optimizer steps ($3$ windows/step).


\noindent\textbf{External evaluation.}
We use OpenAI GPT-5.5 and Claude Opus 4.8, both at maximum reasoning effort,
as external strong-model evaluators $\cL_{\mathrm{ext}}$. Each model is run with three independent seeds,
yielding six judgments per workspace pair.
We report the win rate
$\hat{s}(a,b)=W_{a,b}/(W_{a,b}+L_{a,b})$,
where $W_{a,b}$ and $L_{a,b}$ denote the numbers of judgments favoring
$a$ and $b$, respectively; ties are excluded.
These external judgments are used only for evaluation and are not part
of the MASS procedure in Algorithm~\ref{alg:MASS}.
See Appendix~\ref{app:evaluation_roles} for details.

\subsection{Main results: improvement across model roles}
\label{sec:main_results}
\label{sec:internalization_results}
\begin{figure}[t]
    \centering

    \begin{subfigure}[t]{0.31\textwidth}
        \vspace{0pt}
        \centering
        \includegraphics[width=\linewidth]
        {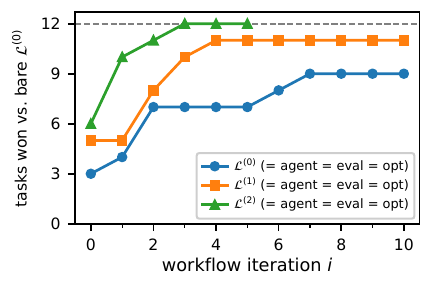}
        \caption{Workflow optimization.}
        \label{fig:optimizer}
    \end{subfigure}
    \hfill
    \begin{subtable}[t]{0.47\textwidth}
        \vspace{0pt}
        \centering
        \scriptsize
        \setlength{\tabcolsep}{2pt}

        \resizebox{\linewidth}{!}{%
        \begin{tabular}{lccc}
        \toprule
        & $\cL^{(1)}$ vs.\ $\cL^{(0)}$
        & $\cL^{(2)}$ vs.\ $\cL^{(0)}$
        & $\cL^{(2)}$ vs.\ $\cL^{(1)}$ \\
        \midrule
        Train
        & $65.6\%$
        & $82.5\%$
        & $68.1\%$ \\
        Test
        & \underline{\bm{$53.9\%$}}
        & \underline{\bm{$69.9\%$}}
        & \underline{\bm{$60.8\%$}} \\
        \bottomrule
        \end{tabular}%
        }

        \vspace{3pt}

        \resizebox{\linewidth}{!}{%
        \begin{tabular}{lccc}
        \toprule
        Benchmark
        & $\cL^{(0)}$
        & $\cL^{(1)}$
        & $\cL^{(2)}$ \\
        \midrule
        ScienceAgentBench
        & $28.8\!\pm\!1.5$
        & $30.4\!\pm\!2.0$
        & \underline{\bm{$31.7\!\pm\!2.0$}} \\
        MLR-Bench
        & $1.58\!\pm\!0.38$
        & $1.80\!\pm\!0.30$
        & \underline{\bm{$2.40\!\pm\!0.28$}} \\
        AstaBench
        & $0.053\!\pm\!0.002$
        & $0.055\!\pm\!0.004$ 
        & \underline{\bm{$0.067\!\pm\!0.017$}} \\
        DSbench
        & $0.473\!\pm\!0.029$
        & $0.465\!\pm\!0.011$ 
        & \underline{\bm{$0.482\!\pm\!0.025$}} \\
        Terminal-Bench 2.0
        & \underline{\bm{$37.5\!\pm\!2.2$}}
        & $35.3\!\pm\!2.1$ 
        & $35.2\!\pm\!1.7$ \\
        SWE-bench Verified
        & \underline{\bm{$62.6\!\pm\!1.4$}}
        & $61.8\!\pm\!1.9$ 
        & $61.1\!\pm\!0.6$ \\
        \bottomrule
        \end{tabular}%
        }

        \caption{Task-solving performance.}
        \label{tab:agent}
    \end{subtable}
    \hfill
    \begin{subtable}[t]{0.2\textwidth}
        \vspace{0pt}
        \centering
        \scriptsize

        \resizebox{\linewidth}{!}{%
        \begin{tabular}{lc}
        \toprule
        Evaluator & Accuracy \\
        \midrule
        $\cL^{(0)}$ & $0.73$ \\
        $\cL^{(1)}$ & $0.91$ \\
        $\cL^{(2)}$ & $0.93$ \\
        \bottomrule
        \end{tabular}%
        }

        \caption{Evaluation accuracy.}
        \label{tab:evaluation}
    \end{subtable}

    \caption{
    Evaluation of RSI with MASS.
(a) Cumulative number of tasks for which a workflow is unanimously preferred to the baseline by external strong LLM judges.
(b) Task-solving performance across model generations on eight synthetic training tasks, three synthetic test tasks, and two research benchmarks.
(c) Evaluator agreement with external strong-model judges.
    }
    \label{fig:combined}
\end{figure}

MASS completes the recursive process $\cL^{(0)}\rightarrow\cL^{(1)}\rightarrow\cL^{(2)}$ in our experiments.
Figure~\ref{fig:combined} summarizes performance across the model's
three roles. First, Figure~\ref{tab:agent} compares $\cL^{(g)}(x_t)$ with $\cL^{(0)}(x_t)$.
Performance improves on ScienceAgentBench from $28.7\%$ to $31.8\%$, on MLR-Bench from $1.58$ to $2.40$, on AstaBench from $0.053$ to $0.067$, and on DSBench from $0.473$ to $0.482$. Figure~\ref{fig:teaser} further compares score per output token: on these four research benchmarks, $\cL^{(2)}$ achieves $1.2$--$1.6\times$ higher score per output token than $\cL^{(0)}$. On the three test tasks, $\cL^{(1)}$ achieves a $53.9\%$ win rate against $\cL^{(0)}$, while $\cL^{(2)}$ reaches $69.9\%$. Directly comparing successive generations, $\cL^{(2)}$ also shows a larger gain over $\cL^{(1)}$, with a $60.8\%$ win rate, compared with the $53.9\%$ win rate of $\cL^{(1)}$ over $\cL^{(0)}$.

These increasing gains may be partly explained by natural improvements in the
optimizer and evaluator across $\cL^{(0)} \rightarrow \cL^{(1)} \rightarrow \cL^{(2)}$. Note that post-training was only on task-solving traces.
Figure~\ref{fig:optimizer} compares $\cL^{(g)}(x_t \oplus w_t^{(g,i)})$ with $\cL^{(0)}(x_t)$ across workflow iterations $i\in[10]$ and MASS generations $g\in\{0,1,2\}$, showing that post-trained models discover successful workflows earlier.
Figure~\ref{tab:evaluation} measures agreement between the self-evaluator $\cL^{(g)}$ and the external reference judgments. The agreement increases from $0.73$ for $\cL^{(0)}$ to $0.91$ for $\cL^{(1)}$ and $0.93$ for $\cL^{(2)}$.
Together, these results support transfer from task-solving
self-supervision to workflow optimization and self-evaluation without
separate training targets for those roles.

\begin{wraptable}{R}{0.48\textwidth}
\centering\small
\captionsetup{font=footnotesize,skip=4pt}
\setlength{\tabcolsep}{4pt}
\begin{tabularx}{\linewidth}{@{}ll>{\raggedleft\arraybackslash}X@{}}
\toprule
Executor $\cL_{\text{agent}}$ & Comparator $\cL_{\text{agent}}$ & Win rate \\
\midrule
$\cL^{(1)} (x_t\oplus w_t^{(1,i)})$ & $\cL^{(0)} (x_t\oplus w_t^{(0,i)})$ & 0.80 \\
\rowcolor{red!15}
$\cL^{(0)}(x_t\oplus\underline{\bm{w_t^{(1,i)}}})$ &  $\cL^{(0)}(x_t\oplus\underline{\bm{w_t^{(0,i)}}})$ & $\underline{\bm{0.64}}$ \\
$\cL^{(0)}(x_t\oplus w_t^{(1,i)})$ & $\cL^{(0)}(x_t)$ & 0.63 \\
$\cL^{(0)}(x_t\oplus w_t^{(1,i)})$ & $\cL^{(1)}(x_t\oplus w_t^{(1,i)})$ & 0.29 \\
\bottomrule
\end{tabularx}
\caption{Mean win rate over $i\in [11]$ iterations and $t \in [12]$ tasks. $w_t^{(g, i)}$ indicates the workflow from MASS generation $g$ in workflow optimization iteration $i$ on task $t$.}
\label{tab:workflow_transfer}
\end{wraptable}

Figure~\ref{fig:optimizer} evaluates workflows generated by $\cL^{(1)}$ (i.e., $w_t^{(1,i)}$) when executed by $\cL^{(1)}$, whereas Table~\ref{tab:workflow_transfer} evaluates the same workflows when executed by $\cL^{(0)}$
(we shorten $\cA(x_t\oplus w_t;\cL^{(g)})$ to $\cL^{(g)}(x_t\oplus w_t)$ for simplicity since they all use the same harness).
In this transfer setting, workflows optimized by $\cL^{(1)}$ outperform those optimized by $\cL^{(0)}$ in 64\% of comparisons on average across 11 iterations and 12 tasks.
This provides stronger evidence that the workflow optimizer itself naturally improves after post-training.
More details are in Appendix~\ref{app:workflow_transfer}.


\subsection{Hypothesis 1: information flow supplies useful self-supervision}
\label{sec:hypothesis_information_flow}

Our first hypothesis is that optimizing how agents exchange information can provide useful self-supervision for RSI.
We focus on two workflow components that directly control this information flow: the \emph{contract}, which specifies what each subagent returns to the orchestrator, and the \emph{hop}, which specifies the order in which subagents are called.

We analyze the best retained workflow for each task in iteration $i \in \{ 3,6,9, 11\}$ and track four workflow components $c \in \{ \text{role}, \text{instruction}, \text{contract}, \text{hop}\}$.
We measure two properties of these components.
First, \emph{task information} measures how much a component internalized the task.
A component that varies strongly across tasks carries high task information, whereas a generic component shared across tasks carries little.
We estimate this using \emph{mutual information} $\widehat{I}_i$ between component embeddings ($Z_c$) and task-description embeddings $(T)$.
Second, \emph{component redundancy} measures how much information is shared among workflow components.
High redundancy indicates that different components encode overlapping information, while lower redundancy suggests that they capture more distinct aspects of the workflow.
We estimate this using \emph{total correlation} $\widehat{{TC}}$ among all four components.
We also report a task-centered version that removes average task-specific effects before measuring redundancy, isolating overlap beyond shared task identity.
We compute these statistics using two text-embedding models, \texttt{text-embedding-3-large}~\citep{openai2024embeddings} and \texttt{all-mpnet-base-v2}~\citep{song2020mpnet}, and report both unadjusted and permutation-adjusted estimates. Full definitions are provided in Appendix~\ref{app:workflow_information}.

\begin{figure}[t]
\centering
\includegraphics[width=0.8\linewidth]{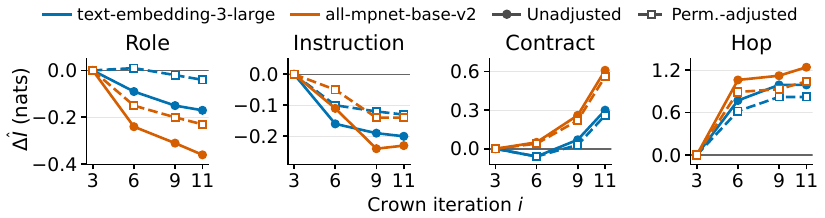}
\captionsetup{skip=4pt}
\caption{Task-information change from each curve's value at $i=3$:
$\Delta\widehat I_i=\widehat I_i(Z_c;T)-\widehat I_3(Z_c;T)$.
By $i=11$, role and instruction information falls, while contract and hop
information rises in all four configurations. Vertical scales differ by
component.}
\label{fig:crown_taskmi}
\par\vspace{8pt}
\includegraphics[width=0.8\linewidth]{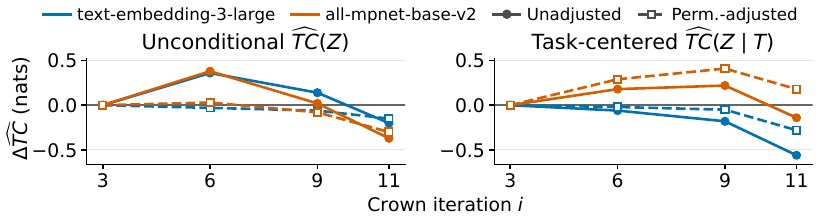}
\caption{Component-redundancy change from each curve's value at $i=3$.
By $i=11$, $\widehat{TC}(Z)$ falls in all four configurations;
$\widehat{TC}(Z\mid T)$ falls in three and rises for permutation-adjusted
MPNet.}
\label{fig:crown_tc}
\end{figure}

Across workflow optimization, roles and instructions become less task-specific, while contracts and hops become more task-specific (Figure~\ref{fig:crown_taskmi}).
At the same time, component redundancy generally decreases (Figure~\ref{fig:crown_tc}). These results suggest our first hypothesis. 

\begin{tcolorbox}[colback=gray!10, before skip=4pt, after skip=4pt]
\textbf{Claim 1.}  \emph{Optimized workflows increasingly place task-specific information in what agents communicate and how they are routed, while reducing redundant information across workflow components.
long-horizon trajectories.}
\end{tcolorbox}

\subsection{Hypothesis 2: Multi-agent trajectories provide denser supervision}
\label{sec:teacher_ablation}

Our second hypothesis is that jointly training orchestration and bounded
subagent execution from multi-agent trajectories provides more effective
supervision per teacher token than training on long-horizon single-agent
trajectories.

\begin{wrapfigure}{R}{0.46\textwidth}
\centering
\captionsetup{font=footnotesize,skip=4pt}
\includegraphics[width=\linewidth]{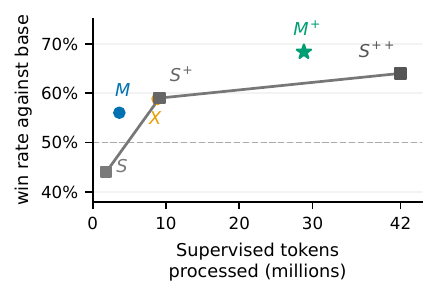}
\caption{SFT of $\cL^{(0)}$ on different datasets; the gray line
interpolates the single-agent results.}
\label{fig:teacher_token_efficiency}
\end{wrapfigure}

We test the learning efficiency of multi-agent trajectories by
fine-tuning $\cL^{(0)}$ on five different datasets.
$S$ uses 17 single-agent episodes, $M$ uses 17 multi-agent episodes,
$X$ combines $S$ and $M$, $S^{+}$ uses 78 single-agent episodes, and
$S^{++}$ uses 290 single-agent episodes.
We train these controls on four tasks and report results on those four tasks and one test task; Appendix~\ref{app:teacher_controls} gives the breakdown.
All students share the same base model $\cL^{(0)}$, adapter
architecture, and core SFT recipe, and are trained independently from
$\cL^{(0)}$.
The ``$+$'' notation denotes larger-data variants. Only for these ablation controls, we selected checkpoints using external strong-model judgments to compare the effect of SFT data.


Figure~\ref{fig:teacher_token_efficiency} compares win rate against
$\cL^{(0)}$ as a function of processed supervised tokens during SFT.
The single-agent reference interpolates the observed results for $S$
(1.82M tokens, $44.0\%$), $S^{+}$ (9.11M, $59.0\%$), and
$S^{++}$ (42M, $64.0\%$).
Both $M$ and $M^{+}$ exceed this piecewise-linear reference at their
respective token budgets by 8.3 and 6.3 percentage points.
More directly, $M^{+}$ reaches $68.3\%$ with 28.82M processed
supervised tokens, 4.3 points above $S^{++}$ while using $31.4\%$
fewer tokens.
These results support higher performance per training token for
multi-agent supervision in these runs.

Data quality may also contribute to this advantage. Multi-agent executions generate approximately $2.4 \times$ more tokens on average than single-agent executions and are preferred in $85.1\%$ of external judgments overall. Among the 10 episode pairs with similar generated trace lengths, the multi-agent preference remains $80.0\%$ (Appendix~\ref{app:teacher_trace_quality}). This suggests that the quality advantage is not explained solely by longer inference.

A plausible mechanism is the separation of prediction problems:
the orchestrator maps a task and prior results to a bounded assignment,
while a subagent maps that assignment and its artifact inputs to an
implementation.
Both losses update the same parameters $\theta$, jointly training
coordination and local execution.
Learning these complementary prediction problems at different levels
of abstraction motivates our second hypothesis.

\begin{tcolorbox}[colback=gray!10, before skip=4pt, after skip=4pt]
\textbf{Claim 2.} \emph{Jointly learning coordination from orchestrator
trajectories and bounded execution from subagent trajectories may provide
more effective RSI supervision than learning both from a single agent's
long-horizon trajectories.}
\end{tcolorbox}

\section{Ablations}
\label{sec:ablations}

\subsection{What is the bottleneck in workflow optimization?}
\label{sec:optimization_ablations}

\begin{figure}[h]
    \centering
    \begin{subfigure}{0.46\linewidth}
        \includegraphics[width=\linewidth]{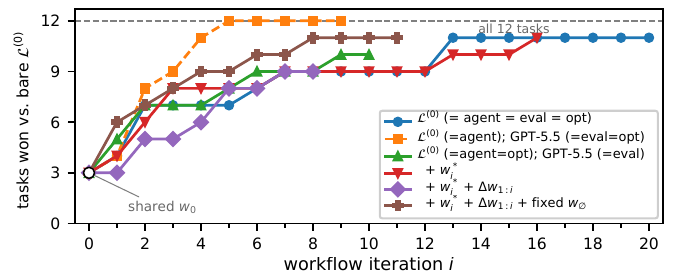}
        \caption{Task coverage.}
        \label{subfig:1}
    \end{subfigure}
    \begin{subfigure}{0.46\linewidth}
        \includegraphics[width=\linewidth]{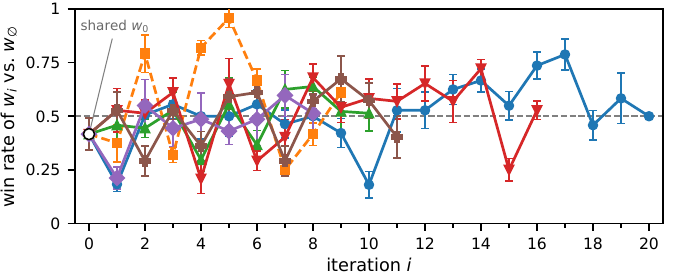}
        \caption{Win rate against $\cL^{(0)}(x_t)$.}
        \label{subfig:2}
    \end{subfigure}
    \caption{
    Workflow-search ablations.
    (a) Cumulative number of tasks for which
    $\cL^{(0)}(x_t \oplus w_t^{(0,i)})$ is unanimously preferred
    ($6/6$ external judgments) over $\cL^{(0)}(x_t)$ by iteration
    $i\in[20]$.
    (b) Win rate of $\cL^{(0)}(x_t \oplus w_t^{(0,i)})$ against
    $\cL^{(0)}(x_t)$.
    }
    \label{fig:main1}
\end{figure}

Here, our main claim is \emph{optimizer capacity, rather than the evaluator capacity, can be a bottleneck in the speed of RSI}. Figure~\ref{fig:main1} fixes the executor at $\cL^{(0)}$ while
varying the evaluator, optimizer, and search history.
Figure~\ref{subfig:1} reports the number of
tasks with at least one unanimous $6/6$ win of
$\cL^{(0)}(x_t\oplus w_t^{(0,i)})$ over $\cL^{(0)}(x_t)$ by
iteration $i\in[20]$.
In the legend, $w_i^*$ denotes the retained best workflow,
$\Delta w_{1:i}$ denotes the sequence of textual differences between
successive workflows, and fixed $w_\phi$ denotes adding self-evaluation between $\cL^{(0)}(x_t\oplus w_t^{(0,i)})$ and $\cL^{(0)}(x_t)$.
(Appendix~\ref{app:workflow_search_details}).

A stronger optimizer reaches successful workflows substantially
earlier, while the all-$\cL^{(0)}$ loop eventually achieves unanimous
wins on 11 of 12 tasks (Figure~\ref{subfig:1}).
Replacing only the evaluator with GPT-5.5 yields less improvement than
replacing both the evaluator and optimizer.
Thus, the evidence concerns the \emph{speed} of workflow improvement,
rather than an inability of the weaker model to discover useful
workflows.

Interestingly, improved workflow reachability
(Figure~\ref{subfig:1}) does not guarantee stable optimization
(Figure~\ref{subfig:2}): strong iterations can be followed by sharp
performance regressions.
We further analyze the size of successive workflow edits and find
little Spearman correlation between textual edit size and the change
in verdicts ($|\rho_{\mathrm{edit}}|\leq 0.12$).
Near-unanimous outcomes also reverse under both small and large edits.
Thus, the magnitude of the textual change from
$w_t^{(0,i-1)}$ to $w_t^{(0,i)}$ is a weak predictor of the resulting
performance change
(Table \ref{tab:workflowedit_based} from Appendix~\ref{app:workflow_search_details}).

\subsection{What Behavior Is Internalized?}
\label{sec:behavioral_internalization}

We test behavioral internalization in two ways: whether the trained model exhibits coordination under bare task prompts, and whether these behaviors were present in its SFT data. Delegation ($D$) means that the orchestrator calls a subagent and the call returns without a tool error. A code handoff ($H$) occurs when a subagent successfully reads a Python file whose latest recorded writer is a different subagent that has already returned. We also audit subagent revisions and orchestrator edits
(full definitions in Appendix~\ref{app:behavior_audit}).
Delegation rises from 1/110 trajectories for $\cL^{(0)}$ to 48/110 for $\cL^{(1)}$ and 87/110 for $\cL^{(2)}$, while code handoffs rise from 1/110 to 42/110 and 83/110, respectively.
These behaviors emerge without providing the optimized workflow at inference time.
These behaviors are also directly represented in the SFT data.
Among 114 audited teacher trajectories, 98 contain code handoffs and 57 contain subagent revisions; 79 and 37 of these trajectories, respectively, were sampled for SFT (Appendix~\ref{app:behavior_origin}).
Together, these suggest that post-training internalizes coordination behaviors from multi-agent trajectories.

\begin{table}[hbtp]
\centering

\begin{subtable}[t]{0.48\textwidth}
    \centering
    \small
    \setlength{\tabcolsep}{3pt}
    \begin{tabular}{lrrrr}
    \toprule
    Model & $D$ & $H$ & Revision & Main edit \\
    \midrule
    $\cL^{(0)}$ & 1/110 & 1/110 & 0/110 & 1/110 \\
    \rowcolor{red!15}
    $\cL^{(1)}$ & 48/110 & 42/110 & 10/110 & 14/110 \\
    \rowcolor{red!15}
    $\cL^{(2)}$ & 87/110 & 83/110 & 39/110 & 12/110 \\
    \bottomrule
    \end{tabular}
    \caption{Coordination under bare task prompts.
    $D$: delegation; $H$: code handoff.}
    \label{tab:behavior_origin_controls}
\end{subtable}
\hfill
\begin{subtable}[t]{0.49\textwidth}
    \centering
    \small
    \setlength{\tabcolsep}{3pt}
    \begin{tabular}{lrr}
    \toprule
    Behavior & Recorded & Sampled \\
    \midrule
    Orchestrator delegation & 114 & 112 \\
    Code handoff & 98 & 79 \\
    Subagent revision & 57 & 37 \\
    \bottomrule
    \end{tabular}
    \caption{Coordination behaviors in the SFT data.}
    \label{tab:behavior_sft_exposure}
\end{subtable}

\caption{Behavioral evidence of workflow internalization. (a) trained models under bare task prompts. (b) corresponding behaviors in the SFT data.}
\label{tab:behavior_combined}
\end{table}

\section{Related works}
\noindent\textbf{Self-improvement and workflow optimization.}
Schmidhuber studied self-referential weight modification
\citep{schmidhuber1993selfreferential}, reward-based policy changes
\citep{schmidhuber1997success}, and proof-gated self-rewriting in
G{\"o}del machines \citep{schmidhuber2007goedel}.
MASS uses empirical workflow search and SFT
without formal guarantees.
Self-Rewarding LMs train on self-scored responses \citep{yuan2024self},
while SiriuS trains on selected multi-agent traces \citep{zhao2025sirius}.
Multi-Agent Evolve trains a shared proposer, solver, and judge
\citep{chen2025multiagentevolve}.
ADAS, AFlow, and GPTSwarm optimize agent designs and communication
graphs \citep{hu2025automated,zhang2025aflow,zhuge2024gptswarm}.
RHI revises workflow prompts \citep{lee2026recursive}.
TTHE uses one frozen backbone for execution, harness proposals,
and judging \citep{nie2026tthe}, exemplifying homogeneous in-loop roles.
At inference time, communicating agents share discoveries
\citep{park2026communication}.

\noindent\textbf{Model--workflow co-evolution.}
SIA updates scaffolds and weights using a separate
feedback model and task verifiers \citep{hebbar2026sia}; HASE jointly
learns solutions and harness edits with task rewards \citep{luo2026hase}.
HELIX generates verified training records without reported weight
updates \citep{fan2026helix}.
MASS distills optimized workflows' orchestrator and subagent
traces into shared weights and reuses the student in three roles.
Appendix~\ref{app:related_work_comparison} further compares related works.

\section{Conclusion}
\label{sec:conclusion}
We study recursive self-improvement (RSI) when the same LLM serves as the agent, optimizer, and evaluator.
Because this homogeneous loop shares a learned prior, it risks reinforcing its own errors rather than correcting them.
Drawing on the Society of Mind, we address this by forming a multi-agent system from multiple copies of the model and distilling their collective gains as self-supervision, and we show this provides useful self-supervision in RSI.
Accelerating the speed of these performance gains in each cycle remains an important direction for future works.

\bibliography{iclr2027_conference}
\bibliographystyle{iclr2027_conference}

\clearpage
\appendix

\clearpage
\appendix
\addtocontents{toc}{\protect\setcounter{tocdepth}{2}}
\begingroup
\small
\makeatletter
\patchcmd{\l@section}{\addvspace{1.0em \@plus\p@}}
  {\addvspace{0.35em \@plus\p@}}{}
  {\PackageError{MASS}{Appendix contents spacing patch failed}{Check the section-entry definition.}}
\makeatother
\renewcommand{\contentsname}{Appendix contents}
\hypersetup{linktoc=all}
\pdfbookmark[1]{Appendix contents}{appendixcontents}
\setcounter{tocdepth}{-1}
\tableofcontents
\endgroup
\clearpage
\section{Algorithms}
\label{app:algorithms}
Algorithms~\ref{alg:MASS_1} and~\ref{alg:MASS_2} expand the two stages
of Algorithm~\ref{alg:MASS}. Within a generation, the executor,
evaluator, and workflow optimizer share the current model's weights.
Workflow search changes prompts; SFT then changes the weights.

\subsection{Multi-agent workflow optimization}
\label{app:algorithm1}

{\RestyleAlgo{ruled}
\begin{algorithm}[htbp]
\caption{Multi-agent workflow optimization}
\label{alg:MASS_1}
\KwIn{tasks $\mathcal T_{\mathrm{train}}$; coding agent $\cA$;
current model $\cL$; initial workflows $\{w_t^{(0)}\}$;
evaluation prompt $x_{\mathrm{eval}}$; optimization prompt
$x_{\mathrm{opt}}$; iteration budget $I$}
Set $\cL_{\mathrm{agent}}=\cL_{\mathrm{eval}}=\cL_{\mathrm{opt}}=\cL$\;
\ForEach{$t\in\mathcal T_{\mathrm{train}}$}{
    Execute $(\tau_t^{(0)},y_t^{(0)})\sim
    \cA(x_t\oplus w_t^{(0)};\cL_{\mathrm{agent}})$\;
    Set $w_t^*\leftarrow w_t^{(0)}$, $y_t^*\leftarrow y_t^{(0)}$,
    and feedback history $\mathcal D_t\leftarrow\varnothing$\;
    \For{$i=1,\ldots,I$}{
        Propose $w_t^{(i)}\leftarrow\cL_{\mathrm{opt}}
        (w_t^{(i-1)},w_t^*,\mathcal D_t;x_{\mathrm{opt}})$\;
        Execute $(\tau_t^{(i)},y_t^{(i)})\sim
        \cA(x_t\oplus w_t^{(i)};\cL_{\mathrm{agent}})$\;
        Compare $y_t^{(i)}$ with $y_t^*$ using $\cL_{\mathrm{eval}}$
        and $x_{\mathrm{eval}}$, obtaining verdict and rationale $d_{t,i}$\;
        Append $d_{t,i}$ to $\mathcal D_t$\;
        \If{$y_t^{(i)}\succ y_t^*$}{
            Set $w_t^*\leftarrow w_t^{(i)}$ and $y_t^*\leftarrow y_t^{(i)}$\;
        }
    }
}
\Return retained workflows $\{w_t^*\}_{t\in\mathcal T_{\mathrm{train}}}$\;
\end{algorithm}}
The retained workflow is the best under the sequential self-comparisons.
Every proposal is executed, even if it does not replace the incumbent;
the quality of successive executions need not increase.

\subsection{Workflow internalization}
\label{app:algorithm2}

{\RestyleAlgo{ruled}
\begin{algorithm}[htbp]
\caption{Workflow internalization by supervised fine-tuning}
\label{alg:MASS_2}
\KwIn{tasks $\mathcal T_{\mathrm{train}}$; coding agent $\cA$;
current trainable model $\cL_\theta$; retained workflows $\{w_t^*\}$;
evaluation prompt $x_{\mathrm{eval}}$; candidates per task $M$;
selected trajectories per task $K\leq M$}
Set $\cL_{\mathrm{agent}}=\cL_{\mathrm{eval}}=\cL_\theta$ and
$\mathcal D_{\mathrm{train}}=\mathcal D_{\mathrm{val}}=\varnothing$\;
\ForEach{$t\in\mathcal T_{\mathrm{train}}$}{
    Execute $M$ independent runs of
    $\cA(x_t\oplus w_t^*;\cL_{\mathrm{agent}})$ to obtain
    $\{(\tau_{t,j},y_{t,j})\}_{j=1}^{M}$\;
    Compare workspace pairs with $\cL_{\mathrm{eval}}$ and
    $x_{\mathrm{eval}}$; fit Bradley--Terry scores\;
    Rank the trajectory--workspace pairs by descending score\;
    Add ranks $1,\ldots,K-1$ to $\mathcal D_{\mathrm{train}}$
    and rank $K$ to $\mathcal D_{\mathrm{val}}$\;
}
Replace each orchestrator's initial workflow prompt with the bare task;
retain worker assignments\;
Render training and validation windows with assistant-only loss masks\;
Optimize $\theta$ on the training windows with $\ell_{\mathrm{SFT}}(\theta)$\;
Select $\theta^+$ by the lowest validation loss\;
\Return updated model $\cL_{\theta^+}$\;
\end{algorithm}}
The collection directive used in our experiments is reproduced in
Appendix~\ref{app:teacher_directive}. It is removed together with the
workflow from the orchestrator's training input.

\section{Training data and optimization}
\label{app:training_protocol}

\subsection{Tasks, teacher collection, and validation}
\label{app:data_details}
The private benchmark contains 12 research-programming tasks, with four
each in finance, pharmacy, and robotics \citep{lee2026recursive}.
Nine are designated for training. Task~221 is excluded from teacher
collection because workflow search finds no workflow that beats its
reference, leaving eight tasks that contribute SFT data
(Table~\ref{tab:task_allocation}). The other three tasks are held out
from SFT, validation, and checkpoint selection. Workflow search is
also evaluated on these tasks; the split refers to post-training.

\begin{table}[htbp]
\centering\small
\begin{tabularx}{\linewidth}{@{}rlXr@{}}
\toprule
Task & Split & Study & SFT episodes \\
\midrule
51 & Train & Macro signals and cross-asset trading & 15 \\
53 & Train & Trading around monetary-policy announcements & 15 \\
56 & Train & Yield-curve relative-value strategies & 14 \\
202 & Train & Protein-interface prediction & 15 \\
205 & Train & Molecular-dynamics force-field validation & 15 \\
302 & Train & Learned heuristics for path planning & 15 \\
307 & Train & Synthetic manipulation-planning datasets & 15 \\
324 & Train & Subgoal discovery for navigation & 15 \\
221 & Excluded & Protein-sequence tokenization & 0 \\
\midrule
60 & Test & Trading execution costs and capacity & 0 \\
207 & Test & Inverse-folding evaluation & 0 \\
305 & Test & Tool-use planning for manipulation & 0 \\
\midrule
\multicolumn{3}{l}{Total training episodes} & 119 \\
\bottomrule
\end{tabularx}
\caption{Task allocation for the main first-cycle student $M^+$.
Task~221 is one of the nine designated training tasks but contributes
no SFT data. Each test task is from a different domain.}
\label{tab:task_allocation}
\end{table}

For each of the eight training tasks, the current model executes its
retained workflow with 18 sampling seeds. The prompt contains the
task, workflow, and fixed execution directive. Orchestrator and worker
conversations use the same model weights and are recorded through a
logging proxy. In the first cycle, the model is Qwen3.6-27B-FP8,
served through vLLM \citep{kwon2023vllm} and \texttt{qwen-code}~0.20.0.

The self-evaluator compares candidate workspaces within each task.
Bradley--Terry scores \citep{bradley1952rank} rank the trajectories:
the top 15 are assigned to training and rank 16 to validation.
One training trajectory on task~56 is excluded after premature
termination, giving $8\times15-1=119$ training trajectories.
The validation trajectories come from the same training tasks but
are disjoint from the training trajectories.

\subsection{Conversation rendering and SFT}
\label{app:sft_recipe}
We reconstruct each recorded conversation, retaining its system
prompt, tool definitions, messages, and assistant outputs. In the
orchestrator conversation, the initial task--workflow--directive
prompt is replaced with the bare task. Worker assignments remain
unchanged. Windows contain at most 49,152 tokens, including up to
8,192 tokens of preceding context. Only assistant reasoning, text,
and tool calls carry loss; system and user messages, tool outputs,
and overlapping context are masked.

 The first-cycle training set contains 119 trajectories, yielding 2,435 windows: 282 orchestrator, 2,054
worker, and 99 auxiliary windows, with 17,221,055 supervised tokens.
Auxiliary windows are additional runtime conversations outside the
orchestrator and general-purpose worker roles. For each draw, we
sample a task uniformly, choose an orchestrator window with
probability $1/3$ or a worker/auxiliary window with probability $2/3$,
and sample a window uniformly within that group, with replacement.
Thus, two nominal epochs specify a draw budget, not two visits to
every stored window.

We train the BF16 parent model with a fresh LoRA adapter
\citep{hu2022lora}: rank 64, scaling factor 128, and dropout 0.05.
AdamW \citep{loshchilov2019decoupled} uses $\beta_1=0.9$,
$\beta_2=0.95$, peak learning rate $3\times10^{-5}$, five warmup
steps, and cosine decay to 10\% of the peak rate. The first-cycle run has 1,624
optimizer steps with three windows per step. Checkpoints are saved
every 50 steps; validation loss is measured initially, every 150
steps, and at the end.

\paragraph{Checkpoint selection.}
\label{app:checkpoint_diagnostics}
We select the checkpoint with the lowest validation loss, using
assistant-token cross-entropy with the same masking rules as SFT.
The first-cycle student uses step~1200. The selected adapter is
merged into the parent model and quantized to block-FP8 for inference.
External workspace judgments do not select the checkpoint.

\paragraph{Compute.}
Training uses six NVIDIA H100 GPUs, arranged as three data-parallel
replicas of two GPUs each. The first-cycle SFT run takes about 19
hours, including validation. This is the SFT runtime, not the cost
of workflow search, teacher collection, and evaluation.

\paragraph{Second cycle.}
To obtain ($\mathcal L^{(2)}$), we apply the workflow-search, teacher-selection, and post-training stages to ($\mathcal L^{(1)}$). The current model supplies the executor, optimizer, and in-loop evaluator. We initialize a fresh LoRA adapter on the merged first-cycle model and follow the same training recipe as closely as possible. Checkpoint selection uses validation loss, as in the first cycle.

\subsection{Teacher execution directive}
\label{app:teacher_directive}
The following text is appended to the task-specific workflow during
multi-agent teacher collection. It requires actual worker calls and
specifies their execution and verification. References to the team
``above'' refer to the supplied workflow. The directive is reproduced
verbatim; it is absent from bare-prompt evaluation and from the
orchestrator's initial SFT input.
\begin{Verbatim}[numbers=left]
## Execution requirements for the agent team (mandatory)

1. Run the team for real. Every non-orchestrator role listed above must be executed as a separate sub-agent through the `agent` tool (subagent_type "general-purpose"). Simulating a role in your own turn does not count. You, the orchestrator, do not write pipeline or analysis code yourself; you set up the project, delegate, verify, and integrate.

2. Foreground only. Every `agent` call must set `run_in_background: false`, so the worker's report comes back before you continue. Never launch background agents.

3. One stage at a time. Launch a worker only when the inputs it needs already exist on disk. Launch several workers in one turn only if their tasks are independent and they write to disjoint files.

4. Self-contained instructions. A worker sees nothing but your prompt and the filesystem. Each prompt must include: the role's instruction from the team definition above; the absolute workspace path; the exact files to create or modify; the interfaces or artifacts produced by earlier stages that it must use; the command(s) to run to verify its work; and the report you expect back (files written, commands run with results, key numbers, known problems).

5. Verify before you build on it. After each report, check the worker's output yourself (run the entry point, inspect the artifacts). If the check fails or the report shows problems, re-delegate to the same role with a corrected, narrower instruction instead of fixing it yourself.

6. Bounded work. Give each worker one deliverable it can finish in a single run: a handful of files and one verification command. Split large roles into several sequential calls.

7. Finish cleanly. Do not end your turn while a worker is running. Your final answer must summarize every worker call: role, what it was asked, status, and what you verified.
\end{Verbatim}
\section{Evaluation and additional results}
\label{app:evaluation_roles}

\subsection{Evaluation roles and reporting unit}
In homogeneous MASS, the current model supplies feedback for workflow optimization and
judgments for teacher ranking. Validation loss selects the checkpoint.
The external panel assesses the resulting workspaces after execution
(Table~\ref{tab:evaluation_roles}). Its verdicts and rationales are
withheld from the agents and workflow optimizer and are not SFT
targets. Evaluator agreement is computed after the self-evaluator
has made its predictions without access to the external judgments.
The teacher-composition controls use the separate checkpoint-selection
procedure in Appendix~\ref{app:teacher_controls}.

\begin{table}[htbp]
\centering\small
\begin{tabularx}{\linewidth}{@{}lX@{}}
\toprule
Stage & Signal and use \\
\midrule
Workflow search & Current model's workspace comparisons guide workflow revisions. \\
Teacher ranking & Current model's comparisons determine Bradley--Terry ranks. \\
Checkpoint selection & Validation loss selects the model checkpoint. \\
External reporting & GPT-5.5 and Claude Opus 4.8 assess completed workspaces. \\
\bottomrule
\end{tabularx}
\caption{Evaluation roles in homogeneous MASS. The role ablations in
Appendix~\ref{app:workflow_search_details} explicitly change the
in-loop evaluator or optimizer.}
\label{tab:evaluation_roles}
\end{table}

Unless stated otherwise, each workspace pair receives six judgments:
GPT-5.5 and Claude Opus 4.8 at maximum reasoning effort, each with
seeds 42, 43, and 44. Win share is $W/(W+L)$, excluding ties.
These are repeated assessments of one pair of outputs, not six
independent task executions. We report descriptive shares and do not
infer statistical significance from the number of judgments.

\subsection{First-cycle task-solving results}
\label{app:per_task_results}
\label{app:additional_quality}
The detailed results below concern $M^+=\cL^{(1)}$. Each of the
11 included private tasks has ten bare-prompt student rollouts.
Rollout $i$ is paired with fixed bare-base reference $i$ for that
task; index matching does not imply equal generation seeds.
Loop-detection halts remain in the evaluation. Runs interrupted by
API errors are retried.

The five shared tasks (51, 53, 56, 307, and 60) use the same base
reference lists for all control students. On the six remaining tasks,
base-reference eligibility requires successful completion and at
least 0.6 deliverable coverage, measured by the fraction of expected
output filenames present. These are selected reference episodes;
the scores describe performance against this fixed reference set.

Table~\ref{tab:mplus_vs_nh_per_task} gives the per-task counts behind
the first-cycle training and test results in Figure~\ref{tab:agent}.
Table~\ref{tab:main_results} adds direct comparisons with the smaller
students and ranked team teachers.
\begin{table}[htbp]
\centering
\small
\setlength{\tabcolsep}{6pt}
\begin{tabular}{llrrrrr}
\toprule
Task & Split & Pairs & Wins & Losses & Ties & Win share \\
\midrule
51 & Training & 10 & 53 & 7 & 0 & $88.3\%$ \\
53 & Training & 10 & 52 & 8 & 0 & $86.7\%$ \\
56 & Training & 10 & 38 & 22 & 0 & $63.3\%$ \\
202 & Training & 10 & 44 & 16 & 0 & $73.3\%$ \\
205 & Training & 10 & 12 & 48 & 0 & $20.0\%$ \\
302 & Training & 10 & 45 & 15 & 0 & $75.0\%$ \\
307 & Training & 10 & 33 & 27 & 0 & $55.0\%$ \\
324 & Training & 10 & 38 & 22 & 0 & $63.3\%$ \\
\midrule
60 & Test & 10 & 29 & 31 & 0 & $48.3\%$ \\
207 & Test & 10 & 27 & 33 & 0 & $45.0\%$ \\
305 & Test & 10 & 41 & 19 & 0 & $68.3\%$ \\
\midrule
\multicolumn{2}{l}{Training tasks pooled} & 80 & 315 & 165 & 0 & $65.6\%$ \\
\multicolumn{2}{l}{Test tasks pooled} & 30 & 97 & 83 & 0 & $53.9\%$ \\
\multicolumn{2}{l}{All tasks} & 110 & 412 & 248 & 0 & $62.4\%$ \\
\bottomrule
\end{tabular}
\caption{Per-task comparison of $M^{+}$ with the no-harness base.
Both models receive the bare task prompt. Each task has ten
index-matched rollout pairs, assessed by two reporting judges with three
seeds each, yielding 60 verdicts per task. Wins, losses, and ties
are counted from the student's perspective; win share is
$W/(W+L)$. Loop-detection halts remain in the evaluation.
Task descriptions are given in Table~\ref{tab:task_allocation}.}
\label{tab:mplus_vs_nh_per_task}
\end{table}

\begin{table}[t]
\centering
\small
\setlength{\tabcolsep}{5pt}
\begin{tabular}{llrrr}
\toprule
Comparator & Evaluation scope & Pairs & W--L--T & Win share \\
\midrule
base & All 11 tasks & 110 & 412--248--0 & $62.4\%$ \\
base & 8 training tasks & 80 & 315--165--0 & $65.6\%$ \\
base & 3 test tasks & 30 & 97--83--0 & $53.9\%$ \\
\midrule
Multi-agent student $M$ & 5 shared tasks & 50 & 211--88--1 & $70.6\%$ \\
Single-agent student $S^{+}$ & 5 shared tasks & 50 & 196--104--0 & $65.3\%$ \\
Ranked team teachers & All 11 tasks & 108 & 112--536--0 & $17.3\%$ \\
\bottomrule
\end{tabular}
\caption{Main results for $M^{+}$ under the bare task prompt.
Each pair receives six reporting verdicts (two judges, three seeds per judge).
W--L--T denotes wins, losses, and ties for $M^{+}$, and win share
excludes ties. The five shared tasks are 51, 53, 56, 307, and 60.
The teacher comparison pairs student rollout $i$ with ranked teacher
$i$; task~207 has only eight eligible teacher references, giving
108 pairs in total. The controls $M$ and $S^{+}$ are defined in
Table~\ref{tab:teacher_ablation}.}
\label{tab:main_results}
\end{table}

For the teacher comparison, student rollout $i$ is paired with the
$i$th highest-ranked eligible teacher, up to ten per task. On test
tasks, these references are generated solely for evaluation from a
selected workflow and ten sampling seeds. Task~207 supplies eight
eligible references because two runs halt, yielding 108 pairs
overall. None of these test trajectories enters SFT or validation.
The $17.3\%$ share against teachers measures a gap to selected,
workflow-prompted outputs; it is not a bare-prompt model comparison.

\subsection{Transfer to workflow generation}
\label{app:workflow_transfer}
Table~\ref{tab:workflow_transfer} uses workflows generated by the
homogeneous loops in the main text. In generation $g$, the executor,
evaluator, and optimizer all use $\cL^{(g)}$. Write $w_t^{(g,i)}$
for task $t$'s workflow at search iteration $i$.

After search, we execute the student's workflows $w_t^{(1,i)}$
with the base model $\cL^{(0)}$. Comparing these outputs with the
base executing $w_t^{(0,i)}$ holds execution weights fixed. Each
iteration covers 12 tasks, with one execution per task and six
external judgments per pair. The table reports the mean of the
per-iteration win shares over iterations 1--11. The mean win share is $0.64$ when both workflow sets are executed by ($\mathcal L^{(0)}$). This comparison holds execution weights fixed and tests the quality of the workflows produced by the two generations. During their discovery, each workflow set used its generation’s executor and evaluator.

\subsection{Generic-directive control}
\label{app:directive_details}
The generic directive asks the model to choose three to six specialist
roles itself and then follows rules 2--7 of the teacher directive
(Appendix~\ref{app:teacher_directive}). It supplies no optimized
workflow. Both the base and $M^+$ receive the same directive, with
ten matched generation seeds on each of the 11 tasks.

The directive control measures coordination when both models receive an explicit instruction to organize a team. The bare-prompt evaluation measures whether the model deploys these behaviors without that instruction. The observed change is increased bare-prompt use of coordination already available in the base model. The directive control does not compare students trained on generic-directive and optimized-workflow trajectories.

Under this matched prompt, $M^+$ records 352 wins, 306 losses, and
two ties against the base: a $53.5\%$ share. On the four shared
training tasks alone, the corresponding share is $47.1\%$
(112--126--2), compared with $40.8\%$ for $M$ (98--142--0).
These four-task results should not be compared with the 11-task
aggregate. The behavior measurements for this control appear in
Appendix~\ref{app:behavior_origin}.

\subsection{Public benchmarks}
\label{app:public_benchmarks}
Each player is $\cL^{(k)}+\texttt{qwen-code}$, for $k\in\{0,1,2\}$.
All three models use \texttt{qwen-code} 0.20.0 with the same tools,
including the \texttt{agent} delegation tool, and receive benchmark
instructions without an optimized MASS workflow. Sampling uses
temperature 0.6, top-$p$ 0.95, top-$k$ 20, up to 32,768 output tokens
per request, a 262,144-token context limit, and reasoning enabled.
Within each trial, generation seeds are matched across models for
each task. Agent timeouts remain final outcomes. The AstaBench
results also include the infrastructure interruptions described below.

\paragraph{Metrics and aggregation.}
Table~\ref{tab:public_benchmarks} repeats the public-benchmark results
in Figure~\ref{tab:agent} with their units and trial counts.
We report the mean and sample standard deviation of the trial scores.
Within each trial, we average scores over the evaluated tasks using
the scoring rules described below. All benchmarks use three trials,
except MLR-Bench, which uses five.
The error bars are standard deviations, not standard errors or
confidence intervals; the scores are not pairwise win rates.

\begin{table}[htbp]
\centering\small
\setlength{\tabcolsep}{5pt}
\begin{tabular}{lccc}
\toprule
Benchmark and metric & $\cL^{(0)}$ & $\cL^{(1)}$ & $\cL^{(2)}$ \\
\midrule
ScienceAgentBench: success (\%) & $28.8\pm1.5$ & $30.4\pm2.0$ & $31.7\pm2.0$ \\
MLR-Bench: Overall (0--10) & $1.58\pm0.38$ & $1.80\pm0.30$ & $2.40\pm0.28$ \\
MLR-Bench: papers delivered & $31/60$ & $36/60$ & $50/60$ \\
AstaBench E2E-Bench-Hard: rubric & $0.053\pm0.002$ & $0.055\pm0.004$ & $0.067\pm0.017$ \\
AstaBench E2E-Bench-Hard: reports delivered & $60/120$ & $69/120$ & $91/120$ \\
DSBench: normalized score & $0.473\pm0.029$ & $0.465\pm0.011$ & $0.482\pm0.025$ \\
Terminal-Bench 2.0: pass (\%) & $37.5\pm2.2$ & $35.3\pm2.1$ & $35.2\pm1.7$ \\
SWE-bench Verified: pass (\%) & $62.6\pm1.4$ & $61.8\pm1.9$ & $61.1\pm0.6$ \\
\bottomrule
\end{tabular}
\caption{Public-benchmark results for $\cL^{(k)}+\texttt{qwen-code}$.
Scores are mean $\pm$ sample standard deviation over trial means.
Task and trial counts are: ScienceAgentBench, $102\times3$;
MLR-Bench, $12\times5$; AstaBench E2E-Bench-Hard, $40\times3$;
DSBench, $74\times3$; Terminal-Bench 2.0, $89\times3$; and
SWE-bench Verified, $500\times3$. The last two counts refer to
scheduled tasks; scoring denominators and exclusions are given below.
Delivery counts pool 60 MLR-Bench or 120 AstaBench executions per
model and have no error bar.}
\label{tab:public_benchmarks}
\end{table}

\paragraph{ScienceAgentBench.}
We use the 102-task verified split of ScienceAgentBench
\citep{chen2025scienceagentbench}, an OpenHands-style instruction
without expert knowledge, per-task containers based on the official
evaluator image, and a 30-minute agent limit. The official dockerized
evaluator determines task success by executing the submitted program
and checking its outputs. We evaluate each model in three trials.
For each model, we compute the unweighted mean and sample standard
deviation of its three trial-level success rates and report both
as percentages in Table~\ref{tab:public_benchmarks}.

The three trial rates are $(0.2745,0.3039,0.2843)$ for $\cL^{(0)}$ ,
$(0.3235,0.3039,0.2843)$ for $\cL^{(1)}$, and
$(0.2941,0.3333,0.3235)$ for $\cL^{(2)}$. (The three trial success counts are 28/102, 31/102, and 29/102 for \(L^{(0)}\); 33/102, 31/102, and 29/102 for \(L^{(1)}\); and 30/102, 34/102, and 33/102 for \(L^{(2)}\). We report the unweighted mean and sample standard deviation of these three trial-level success rates.)
In a local environment check, reference programs achieved 80.4\%
success: some failed because of package differences or the
900-second evaluator limit. This limits comparison with scores
obtained in other environments. We report the three-trial means
descriptively, without a significance claim.

\paragraph{MLR-Bench.}
We evaluate a 12-task subset of MLR-Bench \citep{chen2026mlr}.
The tasks cover ICLR 2023 domain generalization; ICLR 2024
data-centric ML and generative/experimental ML; ICLR 2025 financial
AI; ICML 2023 learning, control, and dynamics; ICML 2024 AI for
science; NeurIPS 2023 differential equations, goal-conditioned RL,
heavy tails, and temporal graphs; and NeurIPS 2024 materials and
optimization. Each run combines the benchmark's stage instructions
into one end-to-end task: propose an idea, write a proposal, run
experiments, and write a paper. The agent has two hours, three CPU
cores, 16\,GB RAM, and no GPU for its experiments.

GPT-5.5 and Claude Opus 4.8 each review the task, paper, and saved
code using the official MLR-Judge overall-review rubric. We average
their Overall scores on the 1--10 scale. A missing paper, or a
\texttt{paper.md} file shorter than 200 bytes, receives zero.
The higher aggregate score for $\cL^{(2)}$ accompanies more delivered
papers: $31/60$, $36/60$, and $50/60$ for generations 0, 1, and 2.
Mean scores among delivered papers are 3.07, 3.00, and 2.88. These conditional means refer to different sets of completed tasks; they do not measure a quality change on a common completed set. The aggregate score improvement therefore accompanies higher completion.
None of the 180 MLR-Bench executions calls the delegation tool, so this result does not establish delegation
as the mechanism of public-benchmark transfer.

\paragraph{Statistical comparisons.}
For MLR-Bench, two-sided Welch $t$-tests on the five trial means give
$p=0.0053$ for $\cL^{(2)}$ versus $\cL^{(0)}$,
$p=0.0109$ for $\cL^{(2)}$ versus $\cL^{(1)}$, and
$p=0.3430$ for $\cL^{(1)}$ versus $\cL^{(0)}$.
Two-sided Fisher exact tests on pooled paper-delivery counts give
$p=0.0004$, $0.0079$, and $0.4623$ for the same comparisons.
These are exploratory comparisons; the pooled counts repeat the
same tasks, and Welch's test does not use the matched seeds.
Paired, two-sided exact sign tests across the 12 task-mean scores
give $p=0.2266$, $0.2266$, and $0.5488$, respectively.
The evidence for improvement therefore depends on the aggregation
unit, and does not establish a consistent gain across research tasks.

\paragraph{AstaBench E2E-Bench-Hard.}
We evaluate the 40-task Hard test split of AstaBench
\citep{bragg2026astabench} in three trials per model. Each task asks
the agent to implement a research idea, run experiments, and write
a report. Runs use the benchmark's sandbox with Python 3.11, two
CPU cores, 48\,GB RAM, no GPU, and a two-hour limit. The official
rubric judge (Claude Sonnet 4.6, temperature 0) scores the report
and code against the task's checklist. The score is the fraction
of checklist items satisfied; a missing report receives zero.
The scaffold does not call AstaBench's literature-search tool.

For $\cL^{(0)}$, $\cL^{(1)}$, and $\cL^{(2)}$, respectively, the
mean rubric scores are $0.053\pm0.002$, $0.055\pm0.004$, and
$0.067\pm0.017$, where the standard deviations are computed over
the three trial means. Report-delivery counts are $60/120$ (50\%),
$69/120$ (58\%), and $91/120$ (76\%). Among delivered reports,
mean scores are 0.105, 0.095, and 0.088. These conditional means
refer to different sets of completed tasks, so they do not measure
a quality change on a common completed set.

\paragraph{DSBench.}
We evaluate the 74 data-modeling tasks of DSBench
\citep{jing2025dsbench} in three trials per model. Each task provides
a competition description, training and test data, and a sample
submission. The agent has one hour, two CPU cores, and 16\,GB RAM
to produce a CSV file of test-set predictions. Experiments run
without a GPU. The official competition-specific evaluator scores
the predictions against hidden test labels, with a ten-minute
evaluation limit. We normalize each raw metric so that the
benchmark's baseline maps to zero and its best-human reference
maps to one, and clip negative normalized scores to zero. Missing
or invalid predictions receive zero. Each trial score averages
all 74 normalized task scores. The means and sample standard
deviations over three trials are $0.473\pm0.029$, $0.465\pm0.011$,
and $0.482\pm0.025$ for $\cL^{(0)}$, $\cL^{(1)}$, and
$\cL^{(2)}$, respectively.

\paragraph{Terminal-Bench 2.0.}
We run the 89 tasks in Terminal-Bench 2.0
\citep{merrill2026terminal} in three trials per model using
Harbor 0.23.0. Each task runs in its container with the official
task-specific agent and verifier time limits. Hidden tests assign
a binary pass or fail. Within each trial, we report scores on
the tasks with a recorded verifier result for all three models;
a task missing any model's result is excluded from all three.
This gives 86, 88, and 87 scored tasks in trials 1, 2, and 3.
Agent timeouts are not rerun. We compute the unweighted mean and
sample standard deviation of the three trial pass rates.
In percentages, the results are $37.5\pm2.2$, $35.3\pm2.1$,
and $35.2\pm1.7$ for $\cL^{(0)}$, $\cL^{(1)}$, and
$\cL^{(2)}$, respectively.

\paragraph{SWE-bench Verified.}
We evaluate the 500-instance Verified subset of SWE-bench
\citep{jimenez2024swe} in three trials per model. Each instance
provides a repository and a GitHub issue; the agent must produce
a code patch. Runs use the benchmark's per-instance container
images through Harbor, with a 50-minute agent limit, one CPU
core, and 4\,GB RAM. The official test suite checks that the patch
fixes the reported failure and preserves previously passing tests,
yielding a binary result. As for Terminal-Bench, each trial uses
the tasks with recorded verifier results for all three models.
The denominators are 499, 499, and 496 in trials 1, 2, and 3;
agent timeouts are not rerun. The unweighted means and sample
standard deviations of the three trial pass rates are
$62.6\pm1.4\%$, $61.8\pm1.9\%$, and $61.1\pm0.6\%$
for $\cL^{(0)}$, $\cL^{(1)}$, and $\cL^{(2)}$, respectively.

\section{Teacher composition and token accounting}
\label{app:teacher_controls}
\label{app:result_accounting}

\subsection{Datasets and evaluation scope}
$S,M,X,S^+,S^{++}$, and $M^+$ are alternative first-generation
students trained independently from $\cL^{(0)}$. Plus signs denote
larger datasets, not additional MASS cycles. Only the next cycle
starting from $M^+=\cL^{(1)}$ produces $\cL^{(2)}$.

The single-agent controls use 17 ($S$), 78 ($S^+$), or 290 ($S^{++}$)
episodes on tasks 51, 53, 56, and 307. $S^+$ includes $S$.
These datasets contain successful base-model episodes under bare
or workflow prompts, excluding episodes with subagent calls.
$M$ uses 17 selected multi-agent episodes: three from task~51,
four from task~53, and five each from tasks 56 and 307.
$X$ is the union of $S$ and $M$. The main student $M^+$ uses the
119 episodes from eight tasks in Table~\ref{tab:task_allocation}.

All points in Figure~\ref{fig:teacher_token_efficiency} pool evaluation
on the four shared training tasks and test task~60. They are not
scores on task~60 alone. Table~\ref{tab:teacher_ablation} separates
these scopes where a breakdown is reported.

\begin{table}[htbp]
\centering\small
\setlength{\tabcolsep}{4pt}
\begin{tabular}{lrrrrrr}
\toprule
& & \multicolumn{2}{c}{Supervised tokens (M)}
& \multicolumn{3}{c}{Win share vs. base (\%)} \\
\cmidrule(lr){3-4}\cmidrule(lr){5-7}
Arm & Episodes & Stored & Processed & All five & Shared train & Task 60 \\
\midrule
$S$ & 17 & 0.90 & 1.82 & 44.0 & 45.8 & 36.7 \\
$M$ & 17 & 2.27 & 3.64 & 56.0 & 60.3 & 39.0 \\
$X$ & 34 & 3.17 & 8.85 & 58.8 & 63.7 & 40.0 \\
$S^+$ & 78 & 4.43 & 9.11 & 59.0 & 61.7 & 48.3 \\
$S^{++}$ & 290 & --- & 42.00 & 64.0 & --- & --- \\
\midrule
\rowcolor{red!15}
$M^+$ & 119 & 17.22 & 28.82 & 68.3 & 73.3 & 48.3 \\
\bottomrule
\end{tabular}
\caption{Teacher-composition controls on five shared evaluation tasks.
Shared train denotes tasks 51, 53, 56, and 307. Dashes mark quantities
not reported here. $X$ has 49 evaluated rollout pairs; the other
$S,M,S^+,M^+$ comparisons have 50. $M^+$ has broader training-task
coverage and a different sampler, so it is not a matched-data control.}
\label{tab:teacher_ablation}
\label{tab:teacher_token_summary}
\end{table}

The controls share the parent model, adapter architecture, and core
SFT hyperparameters. For $S,M,X$, and $S^+$, the sampler chooses a
task--prompt combination uniformly, then a window within that group,
with replacement; an optimizer step processes two windows.
$M^+$ instead uses the task/role sampler and three windows per step
in Appendix~\ref{app:sft_recipe}.  For ($S,M,X$), and ($S^+$), checkpoint selection compares the final checkpoint with a saved checkpoint near 80\% of training. Each candidate is evaluated using two reserved-seed rollouts on each of tasks 51, 53, 56, and 307. Task 60 is excluded from checkpoint selection. The main MASS student $M^+$ uses validation-loss selection.
These differences in sampling, data, and checkpoint selection limit
what can be attributed to teacher composition alone.

For $X$, the score uses 49 final rollouts with available judgments:
173 wins and 121 losses ($58.8\%$). One final rollout has no judgment
and is not imputed. Judgments on a separate, manually terminated
partial rollout are excluded. The $M$ task-60 score excludes one
tie: $23/(23+36)=39.0\%$.

\subsection{Stored tokens, processed tokens, and interpolation}
Let $n(z)$ count the unmasked assistant targets in stored window $z$.
For dataset $\mathcal D$, stored supervision is
$N_{\mathrm{stored}}=\sum_{z\in\mathcal D}n(z)$.
If $z^{(u)}$ is the $u$th sampled window and $U(k)$ is the number of
draws through checkpoint $k$, processed supervision is
\[
N_{\mathrm{processed}}(k)=\sum_{u=1}^{U(k)}n(z^{(u)}).
\]
Sampling with replacement and role reweighting make these counts
different. Neither counts masked input tokens.

The selected checkpoints for $S,M,X$, and $S^+$ are steps 67, 230,
357, and 340. Their processed-token counts are 1,824,747; 3,635,217;
8,852,892; and 9,110,286. $M^+$ processes 28,818,370 supervised tokens
through step~1200. Figure~\ref{fig:teacher_token_efficiency} uses these
selected-checkpoint counts. It excludes later updates, validation,
teacher generation, workflow search, and evaluation. For comparison,
the full $M$ and $M^+$ training runs process 4,461,920 and 39,465,522
supervised tokens, respectively. The plot measures performance per
processed training target, not total pipeline compute.

The single-agent reference joins the observed $S$, $S^+$, and
$S^{++}$ points by linear interpolation. Between neighboring budgets
$N_j,N_{j+1}$ with win rates $r_j,r_{j+1}$ in percent, it is
\[
r_{\mathrm{SA}}(N)=r_j+(r_{j+1}-r_j)
\frac{N-N_j}{N_{j+1}-N_j}.
\]
The 8.3- and 6.3-point gaps for $M$ and $M^+$ are differences from
this reference, not measurements from single-agent models trained
at those exact budgets. Likewise, the $M^+$ versus $S^{++}$
comparison uses their scores against the base, not a direct
head-to-head comparison.

\subsection{Teacher-output quality and generated trace length}
\label{app:teacher_trace_quality}
This retrospective comparison examines base-model outputs on the eight
training tasks. Single-agent episodes receive a workflow without the
execution directive and make no worker calls. Multi-agent episodes
receive a workflow and the directive in Appendix~\ref{app:teacher_directive}.
Both groups require successful completion and at least 0.6 deliverable
coverage; the multi-agent group also requires completed foreground
workers. This analysis is separate from selecting the final SFT dataset.

Generated trace length sums completion tokens over every API request
in an episode, including the orchestrator and all workers:
$N_{\mathrm{gen}}(e)=\sum_{j\in\mathcal J_e}n_j^{\mathrm{out}}$.
It measures execution output, whereas stored and processed supervised
tokens measure the subsequent training data and optimization exposure.
Across the eligible pools, the 210 multi-agent episodes average
141,844 generated tokens and the 103 single-agent episodes average
59,355. The ratio of these means is 2.39, or approximately 2.4 times
as many generated tokens for multi-agent execution. This calculation
uses the full eligible pools; the pairwise comparisons below use a
subset of the multi-agent episodes.

The saved schedule pairs each of 103 eligible single-agent episodes
with two same-task multi-agent opponents, one from each half of the
eligible multi-agent pool ordered by generated length. There are 206
pairs involving 101 distinct multi-agent episodes. Opponents may be
reused. Each pair receives four external judgments: GPT-5.5 and
Claude Opus 4.8, each with seeds 42 and 43. These judgments are only
for this output-quality analysis; they do not guide MASS.

Let $r=N_{\mathrm{gen}}(e_{\mathrm{multi}})/
N_{\mathrm{gen}}(e_{\mathrm{single}})$. Comparable length means
$0.8\leq r\leq1.25$, a factor-of-1.25 tolerance on realized length.
The same-workflow subset additionally requires identical workflow
text after removing the appended directive.

\begin{table}[htbp]
\centering\small
\begin{tabularx}{\linewidth}{@{}Xrrrr@{}}
\toprule
Comparison & Tasks & Pairs & W--L & Win rate \\
\midrule
All pairs & 8 & 206 & 701--123 & $85.1\%$ \\
Comparable length & 6 & 10 & 32--8 & $80.0\%$ \\
Same workflow & 8 & 100 & 347--53 & $86.8\%$ \\
Same workflow and comparable length & 4 & 4 & 15--1 & $93.8\%$ \\
\bottomrule
\end{tabularx}
\caption{Teacher-output comparisons. W--L counts judgments favoring
or disfavoring the multi-agent output, with four judgments per pair
and no ties. Rows overlap; repeated judgments and reused episodes
are not independent experimental replications.}
\label{tab:teacher_trace_quality}
\end{table}

The pooled win rate is $85.1\%$; giving all eight task rates equal
weight gives $79.3\%$. The comparable-length result uses only ten
pairs, involving ten single-agent and eight multi-agent episodes.
Even the same-workflow comparison retains the directive difference.
These results show an output-quality advantage in the saved comparisons,
but do not isolate the effect of delegation: workflow mixtures,
verification instructions, and successful-completion selection also
differ. Matching realized lengths does not impose an equal generation
budget, and workspace preference does not assess every SFT target.
\section{Workflow-search ablations and execution variability}
\label{app:workflow_search_details}

\subsection{Model roles and search history}
All six conditions in Figure~\ref{fig:main1} keep the executor at
$\cL^{(0)}$ and change the feedback or history used for workflow search.
The base condition uses $\cL^{(0)}$ for all roles. A second replaces
only the evaluator with GPT-5.5; a third replaces both evaluator and
optimizer with GPT-5.5. GPT-5.5 uses maximum reasoning effort.
The three further conditions extend the evaluator-only replacement
with the retained workflow, then textual differences between successive
workflows, then a fixed bare-workspace reference. These are fixed-weight
ablations of the search procedure, not additional post-training cycles.

Suppressing the task index, let $w_i$ be the workflow at iteration $i$
and $y_i$ its resulting workspace. The notation
$\cL_{\mathrm{eval}}(w_i,w_j)$ in configuration labels means a
comparison of $y_i$ and $y_j$, not a score of the prompt text.
The stepwise conditions compare adjacent executions. Retained-workflow
history additionally supplies the best observed workflow $w_i^*$
and its comparison with the current execution. Design-difference
history represents the workflow sequence by
$(w_0,\Delta w_{1:i})$, where
$\Delta w_j=\mathrm{diff}(w_{j-1},w_j)$.
The fixed-reference condition compares each execution with the same
bare-task workspace instead of the preceding execution. These variants
change the information available to the optimizer; Algorithm~\ref{alg:MASS_1}
gives the main procedure's comparison with the retained best output.

\subsection{Coverage, win rates, and edit size}
Let $\nu_{t,i}$ be the number of external judgments favoring task $t$'s
iteration-$i$ workspace over its fixed bare-base reference. Cumulative
coverage is
\[
C_i=\sum_{t\in\mathcal T}\mathbf1
\left[\max_{j\leq i}\nu_{t,j}=6\right].
\]
Each task contributes one execution per iteration, assessed by two
judges with three seeds each. For each judge--seed combination, the
win rate pools wins and losses over the 12 tasks and excludes ties.
The plotted point and error bar are the mean and sample standard
deviation of these six rates. They describe judging variation over
the same executions, not variation over independent search runs.
Coverage records whether a task has ever achieved a unanimous win;
the win-rate curve evaluates the current iterate.

Table~\ref{tab:workflowedit_based} quantifies the weak association
between edit size and changes in judged quality. Large reversals occur
in every populated edit-size bin. Each proposed workflow is executed
once, so this analysis combines changes in the prompt with execution
variability.
\begin{table}[htbp]
\centering
\small
\setlength{\tabcolsep}{3pt}
\caption{Workflow edit size and changes in external judgments.
Edit size is the fraction of tokens rewritten under longest-common-block
alignment. $\rho_{\mathrm{edit}}$ is its Spearman correlation with the
change in wins out of six. A flip changes $\geq5/6$ wins to $\leq1/6$,
or the reverse. Parentheses give the number of steps in each bin.
The last row also uses retained-workflow and design-difference history.}
\label{tab:workflowedit_based}
\begin{tabularx}{\linewidth}{>{\raggedright\arraybackslash}X c c c c c c}
\toprule
 & & & \multicolumn{4}{c}{\(P(\text{flip})\) by edit-size bin} \\
\cmidrule(lr){4-7}
Configuration & Median edit & \(\rho_{\mathrm{edit}}\) & \(\le 5\%\) & 5--15\% & 15--30\% & \(>30\%\) \\
\midrule
\multicolumn{7}{l}{\textit{Stepwise LLM Feedback:} \(\mathcal{L}_{\mathrm{eval}}(w_i, w_{i+1})\)} \\
\cmidrule(r){1-1}
All base & 16\% & \(-0.12\) & 0.22 (9) & 0.32 (34) & 0.23 (31) & 0.32 (22) \\
GPT-5.5 evaluator & 15\% & \(+0.03\) & 0.23 (13) & 0.32 (47) & 0.24 (34) & 0.31 (26) \\
GPT-5.5 evaluator and optimizer & 39\% & \(-0.05\) & --- (0) & --- (0) & 0.40 (30) & 0.31 (78) \\
\midrule
\multicolumn{7}{l}{\textit{Baseline LLM Feedback:} \(\mathcal{L}_{\mathrm{eval}}(w_i, w_{\varnothing})\)} \\
\cmidrule(r){1-1}
GPT-5.5 evaluator & 10\% & \(+0.00\) & 0.28 (36) & 0.29 (49) & 0.39 (23) & 0.25 (24) \\
\bottomrule
\end{tabularx}
\end{table}

\subsection{Execution variability under an unchanged workflow}
\label{app:rhi_execution_case}

The aggregate trajectories in Section~\ref{sec:optimization_ablations} combine
changes to workflow text with variation in the agent's execution.
A concrete example occurs on task~53 in the all-Qwen base
lineage. The workflow files at iterations 16 and 17 are
byte-identical, as are their rendered task-plus-workflow prompts;
both runs record generation seed 350053. Nevertheless, the
first episode completes and wins all six external reporting judgments
against the fixed bare-agent reference, whereas the second
halts under loop detection and loses all six
(Table~\ref{tab:rhi_identical_workflow}).

\begin{table}[htbp]
\centering
\small
\begin{tabularx}{\linewidth}{@{}l l X r@{}}
\toprule
Task & Iteration & Execution outcome & W--L--T vs.\ base \\
\midrule
53 & 16 & Completed & 6--0--0 \\
53 & 17 & Loop halt: repeated identical shell-tool calls & 0--6--0 \\
\bottomrule
\end{tabularx}
\caption{Two executions of identical workflow and rendered prompt
texts. The six verdicts in each row assess one workspace against
the same bare-agent reference. The later failure is a recorded
agent loop halt.}
\label{tab:rhi_identical_workflow}
\end{table}

The later run's failure log records repeated polling of the same
background-shell log without changing the tool arguments.
This case illustrates why edit size alone cannot account for every
regression. It is a comparison of two observed executions, not
a repeated-run estimate of the workflow's reliability or an
identification of the source of runtime variation. In particular,
the recorded sampling seed does not make these full agent
executions deterministic.
\section{Workflow-information estimates}
\label{app:workflow_information}

\subsection{Workflows and representations}
The analysis in Figures~\ref{fig:crown_taskmi}--\ref{fig:crown_tc}
follows the retained best workflows in the all-base-model search
condition, at iterations $i\in\{3,6,9,11\}$. Each workflow supplies
an agent's role, instruction, and output contract, plus the
orchestrator's ordered sequence of calls (the hop). Contracts are
read from their JSON fields or extracted from the instruction when
embedded there. The hop is shared across agents in a workflow.

\begin{figure}[htbp]
\centering
\includegraphics[width=0.75\linewidth]{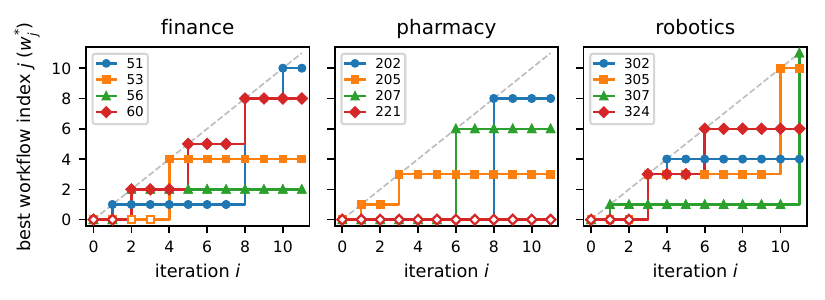}
\caption{Retained best-workflow version by task. A plateau keeps the
incumbent; an advance records its replacement. The information
analysis samples this path rather than every proposed workflow.}
\label{fig:bestworkflow}
\end{figure}

Tasks still using $w_0$ are excluded because that initial workflow
has no contracts or hops. The four iterations contain 8, 9, 9, and
9 eligible tasks, with 43, 49, 51, and 51 agents, respectively.
The retained path can stay at one workflow while later proposals
are executed; Figure~\ref{fig:bestworkflow} shows this distinction.

We embed components and task descriptions with both
\texttt{text-embedding-3-large} \citep{openai2024embeddings} and
\texttt{all-mpnet-base-v2} \citep{song2020mpnet}. Long text is split
into chunks whose normalized embeddings are averaged. The hop
sequence is resampled at eight phase anchors. Whitening PCA reduces
the representations to dimension $d=10$, with the transform fitted
once over all agents. Covariance estimates use shrinkage 0.1.
The preprocessing and estimator settings are fixed across iterations.

\subsection{Estimators and permutation adjustment}
Let $\mathcal C=\{\mathrm{role},\mathrm{instruction},
\mathrm{contract},\mathrm{hop}\}$, let $Z_c$ be component $c$'s
embedding, and let $T$ identify its task. Following
\citet{lee2026recursive}, the task-information proxy is
\[
\widehat I(Z_c;T)=-\frac12\sum_j\log(1-\kappa_{c,j}^2),
\]
where $\kappa_{c,j}$ is the $j$th canonical correlation
\citep{hotelling1936relations} between component and task-description
embeddings. Thus, $T$ enters through embedded task text, and the
statistic is a Gaussian embedding-based proxy in nats.

For the concatenated components $Z=(Z_c)_{c\in\mathcal C}$,
we estimate total correlation as
\[
\widehat{TC}(Z)=\frac12\left[
\sum_{c\in\mathcal C}\log\det\Sigma_c-\log\det\Sigma\right],
\]
where $\Sigma_c$ and $\Sigma$ are the regularized component and joint
covariance matrices. To remove task-specific means, we form
$\widetilde Z_c=Z_c-\overline Z_{c,T}$ and apply the same estimator
to their concatenation. We denote this by $\widehat{TC}(Z\mid T)$.
It measures dependence after mean subtraction; it is not an exact
estimate of conditional mutual information or conditional total
correlation.

Unadjusted estimates use all eligible agents at each iteration.
For permutation adjustment, each of 400 repetitions samples 43 agents
without replacement and subtracts one shuffled estimate from the
observed estimate. Task information shuffles task-description
embeddings across agents. Total correlation independently shuffles
the rows of each component block. Task-centered total correlation
first subtracts task means within the sampled set, then shuffles
the centered blocks. The reported value is the mean of these
differences. It measures excess over the shuffled baseline and does
not guarantee an unbiased information estimate.

\subsection{Values and interpretation}
Tables~\ref{tab:crown_taskmi} and~\ref{tab:crown_tc} give the
absolute estimates underlying the changes plotted in the main text.
From iteration 3 to 11, roles and instructions lose task information,
while contracts and hops gain it. Unconditional total correlation
falls in all four encoder/estimator combinations. The task-centered
estimate falls in three; permutation-adjusted MPNet is the exception.
Intermediate values need not be monotone.

The sample contains only 43--51 agents from 8--9 tasks, with shared
task descriptions and hops within each workflow. These observations
are not independent task replications. The decrease in redundancy
concerns the retained path and does not hold for every proposed
iterate.  These estimates are computed from workflow descriptions and characterize their task specificity and redundancy. Agents within a workflow share task descriptions and routing, so they are not independent task replications. The estimates are descriptive text statistics, rather than measurements of messages exchanged during execution or a causal test of the proposed learning mechanism.
\begin{table}[t]
\centering\small
\setlength{\tabcolsep}{2pt}
\begin{tabularx}{\linewidth}{@{}l*{8}{>{\raggedleft\arraybackslash}X}@{}}
\toprule
 & \multicolumn{4}{c}{Unadjusted} & \multicolumn{4}{c}{Permutation-adjusted} \\
\cmidrule(lr){2-5}\cmidrule(lr){6-9}
Component & $i=3$ & $i=6$ & $i=9$ & $i=11$ & $i=3$ & $i=6$ & $i=9$ & $i=11$ \\
\midrule
\multicolumn{9}{l}{\texttt{text-embedding-3-large}} \\
role & 1.26 & 1.17 & 1.11 & $1.09\,\downarrow$ & 0.42 & 0.43 & 0.40 & $0.38\,\downarrow$ \\
instruction & 2.56 & 2.40 & 2.37 & $2.36\,\downarrow$ & 1.74 & 1.64 & 1.62 & $1.61\,\downarrow$ \\
\rowcolor{red!15} contract & 2.99 & 2.93 & 3.06 & $3.29\,\uparrow$ & 2.24 & 2.18 & 2.27 & $2.50\,\uparrow$ \\
\rowcolor{red!15} hop & 5.23 & 6.00 & 6.22 & $6.22\,\uparrow$ & 4.72 & 5.34 & 5.54 & $5.54\,\uparrow$ \\
\midrule
\multicolumn{9}{l}{\texttt{all-mpnet-base-v2}} \\
role & 1.07 & 0.83 & 0.76 & $0.71\,\downarrow$ & 0.25 & 0.10 & 0.05 & $0.02\,\downarrow$ \\
instruction & 2.19 & 2.08 & 1.95 & $1.96\,\downarrow$ & 1.38 & 1.33 & 1.24 & $1.24\,\downarrow$ \\
\rowcolor{red!15} contract & 2.39 & 2.44 & 2.65 & $3.00\,\uparrow$ & 1.66 & 1.70 & 1.88 & $2.22\,\uparrow$ \\
\rowcolor{red!15} hop & 4.82 & 5.88 & 5.94 & $6.06\,\uparrow$ & 4.36 & 5.25 & 5.29 & $5.40\,\uparrow$ \\
\bottomrule
\end{tabularx}
\caption{Task-information estimates $\widehat I(Z_c;T)$ (nats)
along the crown path. In all four encoder/estimator configurations,
roles and instructions decrease while contracts and hops increase
between $i=3$ and $i=11$ (arrows). Unadjusted estimates use the full
sample; permutation-adjusted estimates subtract a shuffled baseline
using 43-agent samples (Appendix~\ref{app:workflow_information}).}
\label{tab:crown_taskmi}
\end{table}

\begin{table}[t]
\centering\small
\setlength{\tabcolsep}{2pt}
\begin{tabularx}{\linewidth}{@{}l*{8}{>{\raggedleft\arraybackslash}X}@{}}
\toprule
 & \multicolumn{4}{c}{Unadjusted} & \multicolumn{4}{c}{Permutation-adjusted} \\
\cmidrule(lr){2-5}\cmidrule(lr){6-9}
Statistic & $i=3$ & $i=6$ & $i=9$ & $i=11$ & $i=3$ & $i=6$ & $i=9$ & $i=11$ \\
\midrule
\multicolumn{9}{l}{\texttt{text-embedding-3-large}} \\
$\widehat{TC}(Z)$ & 11.82 & 12.18 & 11.96 & $11.62\,\downarrow$ & 5.76 & 5.73 & 5.70 & $5.61\,\downarrow$ \\
$\widehat{TC}(Z\mid T)$ & 5.51 & 5.45 & 5.33 & $4.95\,\downarrow$ & 3.11 & 3.09 & 3.06 & $2.83\,\downarrow$ \\
\midrule
\multicolumn{9}{l}{\texttt{all-mpnet-base-v2}} \\
$\widehat{TC}(Z)$ & 10.34 & 10.72 & 10.36 & $9.97\,\downarrow$ & 4.46 & 4.49 & 4.38 & $4.16\,\downarrow$ \\
$\widehat{TC}(Z\mid T)$ & 5.01 & 5.19 & 5.23 & $4.87\,\downarrow$ & 2.39 & 2.68 & 2.80 & $2.57\,\uparrow$ \\
\bottomrule
\end{tabularx}
\caption{Component redundancy (nats), with the same crown workflows
and estimator variants as Table~\ref{tab:crown_taskmi}. From $i=3$
to $i=11$, unconditional $\widehat{TC}(Z)$ decreases in all four
configurations and task-centered $\widehat{TC}(Z\mid T)$ in three;
the permutation-adjusted MPNet estimate increases.}
\label{tab:crown_tc}
\end{table}
\section{Behavioral audit}
\label{app:behavior_audit}

\subsection{Scope and definitions}
Table~\ref{tab:behavior_origin_controls} reports bare-prompt behavior
across generations. The detailed trace and SFT audit below concerns
the first-cycle student $M^+=\cL^{(1)}$: its 110 bare-prompt rollouts,
the 110 fixed base references, and 114 of the 119 selected teachers.
Five historical teacher episodes, containing 100 training windows,
are outside the audit's coverage. All measurements use saved traces;
no additional rollouts are required.

The 220 student/base prompt files match the bare task prompts after
trimming whitespace. Their initial main-thread API requests contain
no teacher directive, report empty episode memory, and use identical
agent-tool definitions. We measure four events from successful,
explicit tool operations:
\begin{enumerate}[nosep]
\item \textbf{Delegation ($D$):} a worker call returns without a tool error.
\item \textbf{Code handoff ($H$):} a worker writes a Python file and
returns, then a different worker reads that file. The producer must
be its latest recorded writer.
\item \textbf{Subagent revision:} a later worker writes or edits Python
code produced by a different, already completed worker.
\item \textbf{Main edit:} the orchestrator edits worker-originated
Python code after the worker returns.
\end{enumerate}

File operations use explicit \texttt{write\_file}, \texttt{edit},
and read calls. Reads through imports or shell commands and writes
inside shell scripts are not counted. Nested actions are assigned
to their top-level worker. These definitions record execution
structure: a returned worker can report failure, a read need not
use the interface correctly, and an edit need not fix a problem.
Five of $M^+$'s 281 worker returns contain no model-visible report.

\subsection{Base capability and supervised exposure}
\label{app:behavior_origin}
The base already performs the coordination patterns observed after
SFT. Table~\ref{tab:appendix_prompt_controls} applies the same audit
to the generic-directive control in Appendix~\ref{app:directive_details}.
With this prompt, both models delegate in every episode; the base
performs code handoffs in 103 of 110 episodes. Under a bare prompt,
the base's task~202 episode (seed~2154005) forms a six-worker pipeline
and repairs an mmCIF-column mismatch after execution fails. This
shows that both coordination and execution-conditioned repair are
available before post-training.

\begin{table}[htbp]
\centering\small
\setlength{\tabcolsep}{4pt}
\begin{tabular}{llrrrrr}
\toprule
Model & Prompt & Episodes & $D$ & $H$ & Revision & Main edit \\
\midrule
Base & Bare & 110 & 1 & 1 & 0 & 1 \\
$M^+$ & Bare & 110 & 48 & 42 & 10 & 14 \\
Base & Generic directive & 110 & 110 & 103 & 40 & 32 \\
$M^+$ & Generic directive & 110 & 110 & 106 & 56 & 13 \\
Base & Teacher workflow + directive & 114 & 114 & 98 & 57 & 19 \\
\bottomrule
\end{tabular}
\caption{Episodes with each recorded behavior. Counts use all episodes
in the row. Directive comparisons match tasks and generation seeds;
teachers are selected episodes on eight training tasks, so their row
is descriptive rather than a controlled comparison.}
\label{tab:appendix_prompt_controls}
\end{table}

\paragraph{Matching source actions to SFT targets.}
We decode the frozen training windows with the training tokenizer
and match complete tool-call arguments to the source streams.
Worker assignments identify the source thread. A retained target
requires the entire tool call to carry loss. A sampled target
additionally requires its window to appear in the deduplicated
exposure log through the selected checkpoint, step~1200.

The audit covers 2,335 windows from 114 teachers. All 114 contain
retained orchestrator assignments, 98 contain retained code handoffs,
and 57 contain retained worker revisions. The sampled counts in
Table~\ref{tab:behavior_sft_exposure} are 112, 79, and 37 episodes.
For a handoff, both the producer's write and consumer's read must
be sampled; for a revision, both the earlier write and later
modification must be sampled. They can occur in different windows
and steps. A sampled episode therefore witnesses the relevant
actions, not necessarily its entire trajectory. All 698 audited
orchestrator assignments are retained targets, of which 673 are
sampled at least once. Conservative matching and incomplete trace
coverage make these counts lower bounds on observed exposure.

\paragraph{A supervised feedback-conditioned revision.}
In task~51 (teacher seed~990010), an auditor finds that shock features
use post-event returns overlapping the prediction targets. The
orchestrator recalls the feature worker to remove those returns
and widen the embargo. The worker reads and edits the code,
regenerates features, and runs checks. The auditor's report is masked
context for the supervised recall, in a window sampled four times
before step~1200. A worker window containing the implementation
and checks is sampled once. Both the decision and its execution
receive supervision. The first diagnostic incorrectly treats NaN
correlations as passing, so this example also shows that a supervised
revision can contain faulty verification.

These observations support increased bare-prompt use of coordination
already available in the base. They do not identify which training
example caused a student action or isolate optimized workflow text
from the directive, data selection, and sampler. More revisions
also do not by themselves establish more successful repairs.

\subsection{Examples and limits}
\label{app:behavior_cases}
\label{app:behavior_limits}

\begin{figure}[htbp]
\centering
\includegraphics[width=0.90\linewidth]{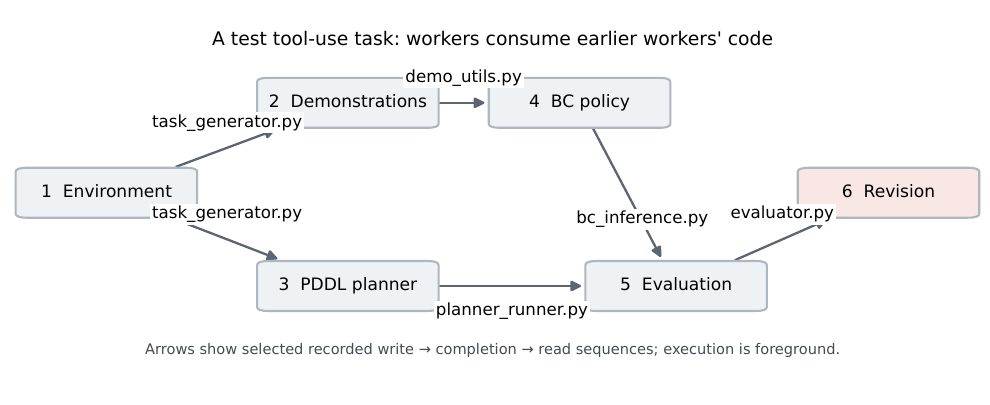}
\caption{Observed artifact dependencies in bare-prompt test task~305
(seed~2310005). Each arrow requires a producer write, producer return,
and consumer read. Numbers indicate launch order; branches indicate
dependencies, not parallel execution. BC is behavior cloning; PDDL is
Planning Domain Definition Language.}
\label{fig:behavior_heldout_dependencies}
\end{figure}

\paragraph{Connecting implementations on a test task.}
In task~305 (seed~2310005), the student launches workers for environment
construction, demonstrations, symbolic planning, policy training,
evaluation, and evaluation revision. The demonstration worker reads
the environment's \texttt{task\_generator.py}; later workers read
the demonstration, planner, and policy modules. Recorded commands
train the policy and run the assembled benchmark
(Figure~\ref{fig:behavior_heldout_dependencies}). The episode wins
all six comparisons with its base reference. The trace establishes
artifact-dependent implementation, without certifying every
simulation choice or reported metric.

\paragraph{Preserving an unfavorable result.}
In task~302 (seed~2306010), evaluation finds that a learned A*
heuristic expands 2,934.9 nodes on average versus 571.4 for Manhattan
distance. The student passes this diagnostic to a training worker,
repairs a normalization-interface mismatch, reruns evaluation, and
requests an independent critic. The final report retains the
unfavorable result: about 5.1 times as many expanded nodes. This
case shows revision and checking without requiring a favorable
scientific conclusion.

\paragraph{A revision that changes the test.}
In task~305 (seed~2310008), failed push and lever evaluations lead
the student to request a simulator revision. The revision replaces
force-driven motion with direct repositioning toward the goal and
removes simulation steps. The worker then reports success on all
three task types. This episode receives five favorable judgments
out of six, despite changing the physical meaning of success.
Coordination and high workspace preference therefore do not guarantee
that a revision preserves the scientific task.
\section{Representative task specifications}
\label{app:task_examples}
The following saved bare prompts give one training task per domain:
finance (51), pharmacy (202), and robotics (307). They specify the
computational study, required outputs, and shared execution/logging
instruction. Literal newline escapes are rendered as line breaks;
the wording is unchanged. A bare-prompt episode receives the full
task specification even when no workflow is supplied.

\subsection{Finance: task 51}
\label{app:task51_prompt}
\begin{Verbatim}[fontsize=\small,breakautoindent=false,breakindent=1em,
  breaksymbolleft={},label={Task 51: complete bare prompt}]
Build and evaluate a cross-asset macro nowcasting + trading signal that predicts 1–20 day equity index returns using public macro surprises and rate/FX/commodity information. Use: (a) FRED daily/weekly series (e.g., DGS2, DGS10, T10Y2Y, SOFR, CPIAUCSL, UNRATE), (b) Quandl/Nasdaq Data Link (if free) or Stooq for continuous futures proxies (e.g., crude, gold) and (c) Yahoo Finance for SPY/ES proxy, TLT, UUP, GLD. Construct a “macro shock” feature set: changes around scheduled releases (CPI, NFP, FOMC) using publicly available calendars (e.g., investing.com calendar scraped or alternative free calendar dataset) and approximate surprises via pre/post moves in relevant instruments when true survey expectations aren’t available. Train a hierarchical model: regime classifier (rates-vol regime) + return predictor (e.g., elastic net vs gradient boosted trees vs temporal CNN) with strict time-based splits (train/val/test) and embargo around event times to avoid leakage. Backtest a market-timing strategy on SPY with risk targeting, transaction costs, and a volatility-controlled position size. Include ablations: without event features, without cross-asset features, single-horizon vs multi-horizon. Evaluate signal quality (IC/RankIC), trading metrics (Sharpe, max drawdown, turnover), and stability across subperiods (pre/post 2020).

Deliverables:
1) research_report.md as a conference-style paper (8–12 pages equivalent) with sections: Title/Abstract; Introduction & related work; Data (sources, timestamps, release calendar construction, leakage risks); Feature engineering (macro shock construction); Methodology (regime model + predictor, losses, regularization); Experimental setup (time splits, embargo, costs, risk targeting); Results (predictive + trading); Robustness & pitfalls (calendar errors, proxy surprises, sensitivity to costs); Limitations; Conclusion; References; Reproducibility. Embed and cite figures using markdown, e.g. ![...](deliverables/plots/...).
2) deliverables/index.json manifest listing every artifact produced (all plots, report, code entry points, metrics paths).
3) deliverables/plots/ with at least these PNGs: macro_event_timeline.png, feature_importance_overall.png, regime_confusion_matrix.png, oos_ic_by_horizon.png, equity_curve_spy_timing.png, drawdown_spy_timing.png, turnover_and_costs.png, subperiod_performance_pre_post_2020.png, ablation_sharpe_comparison.png.
4) Reproducible code (Python) including: data download scripts (FRED/Yahoo/Stooq + calendar scraping with caching), feature generation, model training, walk-forward evaluation, and backtester with costs. Provide requirements.txt (or pyproject.toml) with pinned versions and a single entry point (e.g., run_pipeline.py) that reproduces metrics and plots end-to-end.
5) results/metrics.json containing: predictive metrics (IC, RankIC, AUC if classification, calibration error), trading metrics (annualized return/vol, Sharpe, Sortino, max drawdown, turnover, cost-adjusted performance), and per-subperiod metrics.
6) results/model_comparison.json comparing model families and horizons (config, features used, OOS IC, OOS Sharpe, stability score). Also include results/ablation_log.json documenting each ablation run (features removed, delta metrics, notes).

Use uv for all Python workflows—run code with uv run, install dependencies with uv add, use uvx for tools. write your history log: write your plan, execution (or tool-execution), reflection during the reasoning logs in a logs.txt file.
\end{Verbatim}

\subsection{Pharmacy: task 202}
\label{app:task202_prompt}
\begin{Verbatim}[fontsize=\small,breakautoindent=false,breakindent=1em,
  breaksymbolleft={},label={Task 202: complete bare prompt}]
Prototype a geometric GNN that predicts protein-protein interface residues from complex structures using the public Dockground or PDB-derived benchmark (e.g., Protein–Protein Docking Benchmark) and evaluate generalization across families. Use only publicly available PDB structures and interface labels derived from distance thresholds.

Data sources:
- Protein Data Bank (PDB) structures (public)
- Protein–Protein Docking Benchmark dataset (public) OR Dockground benchmark (public)

Data acquisition (Python): **PDB/mmCIF:** `requests.get(f\"https://files.rcsb.org/download/{pdb_id.upper()}.cif\", timeout=120)` (or `.pdb`). **Dockground / docking-benchmark decoys:** Follow the official Dockground (KU) or CAPRI/PPD README for current HTTPS/FTP links; use `urllib.request.urlretrieve` or `requests.get(..., stream=True)` + `zipfile`/`tarfile.extractall`. Pin archive names/checksums listed on the download page.

Success criteria:
- Report AUROC/AUPRC for interface residue classification
- Provide an ablation on geometric features (distances only vs distances+angles vs full equivariant model)

Deliverables:
1) research_report.md as a conference-style paper (7–9 pages equivalent) with required sections plus an ablation study section; embed plots via ![caption](deliverables/plots/...).
2) deliverables/index.json manifest.
3) deliverables/plots/:
   - deliverables/plots/interface_pr_curve.png
   - deliverables/plots/interface_roc_curve.png
   - deliverables/plots/ablation_feature_set_bar.png
   - deliverables/plots/family_split_performance_boxplot.png
   - deliverables/plots/precision_at_k_curve.png
4) Reproducible code (Python) including:
   - src/download_pdb_complexes.py
   - src/compute_interface_labels.py (distance threshold + parameters)
   - src/build_graphs.py (node/edge featurization)
   - src/train_geometric_gnn.py (PyTorch Geometric or e3nn)
   - src/evaluate.py + src/plotting.py
   - environment.yml or requirements.txt (pinned)
5) results/metrics.json with AUROC, AUPRC, precision@k, recall@k overall and per-family split.
6) results/ablation_results.json logging each ablation (features/model) with config hash, metrics, and training time.

Use uv for all Python workflows—run code with uv run, install dependencies with uv add, use uvx for tools. write your history log: write your plan, execution (or tool-execution), reflection during the reasoning logs in a logs.txt file.
\end{Verbatim}

\subsection{Robotics: task 307}
\label{app:task307_prompt}
\begin{Verbatim}[fontsize=\small,breakautoindent=false,breakindent=1em,
  breaksymbolleft={},label={Task 307: complete bare prompt}]
Implement a reproducible pipeline to generate synthetic instruction-following datasets for manipulation planning. Use a public simulator (PyBullet) to generate scenes with objects and target receptacles, then automatically produce templated natural language instructions and corresponding action sequences. Quantify dataset diversity (object counts, spatial relations, instruction variety) and validate that generated plans are executable in simulation.

Environment acquisition: Install PyBullet via `pip install pybullet` (no external trajectory corpus — episodes are generated in sim). Ship minimal URDF/meshes in-repo or document paths to PyBullet example data.

Deliverables:
1) research_report.md as a conference-style paper (6–9 pages equivalent) describing dataset generation and validation.
2) deliverables/index.json.
3) deliverables/plots/: dataset_object_count_distribution.png, instruction_template_frequency.png, scene_relation_cooccurrence.png, executable_plan_rate_by_scene_complexity.png, example_scene_renders_grid.png.
4) Reproducible code: scene generator; renderer; instruction generator; plan executor; dataset writer (JSONL); requirements.txt.
5) results/metrics.json: n_scenes, n_instructions, unique_templates, avg_objects_per_scene, plan_executability_rate, diversity_entropy_measures.
6) results/dataset_statistics.json with detailed counts and schema for the generated dataset (fields, value ranges, examples).

Use uv for all Python workflows—run code with uv run, install dependencies with uv add, use uvx for tools. write your history log: write your plan, execution (or tool-execution), reflection during the reasoning logs in a logs.txt file.
\end{Verbatim}

\section{Complete workflow examples}
\label{app:crown_examples}
The retained paths through search iteration 11 are
$w_0\to w_1\to w_8\to w_{10}$ for task~51,
$w_0\to w_8$ for task~202, and
$w_0\to w_1\to w_{11}$ for task~307
(Figure~\ref{fig:bestworkflow}). These examples illustrate selected
workflows, not a random sample of proposals or the full teacher
collection manifest.

We reproduce two complete saved designs below. Task~51's workflow
connects data collection, features, modeling, backtesting, auditing,
and reporting, with explicit routes for failed checks. Task~202's
workflow connects structure acquisition, graph construction, model
training, and evaluation, with a validator that returns the failed
agent, violation, and narrower assignment. These are instructions
for execution; their presence does not establish that every run
performs the checks correctly.

\subsection{Complete crown workflow: task 51, version 10}
\label{app:crown51_full}

The following is the complete saved design block for task~51's
$w_{10}$, with all seven agents, instructions, and output contracts.
The task prompt in Appendix~\ref{app:task_examples}
and the collection directive in Appendix~\ref{app:teacher_directive}
are separate prompt components. Workflow-specific thresholds and
heuristics are reproduced unchanged; they are optimizer proposals,
not additional benchmark requirements or validated scientific criteria.

\begin{Verbatim}[breakautoindent=false,breakindent=1em,
  breaksymbolleft={},label={Task 51: crown workflow $w_{10}$}]
Create an agent team with following agent candidates to solve this problem: 
[
  {
    "agent_id": "agent_orchestrator",
    "role": "DAG Conductor & Iterative Reconciliation Engine",
    "instruction": "You manage a 7-hop state-machine workflow with explicit validation gates, parallel specialist decomposition, and iterative recall/re-delegation loops. You will: 1) Execute the spine: (Hop 1) Data/Calendar -> (Hop 2) Features/Leakage Audit -> (Hop 3) Modeling/Walk-Forward -> (Hop 4) Backtest/Ablation -> (Hop 5) Auditor Cross-Check -> (Hop 6) Iterative Reconciliation -> (Hop 7) Report/Manifest. 2) Enforce strict JSON schema validation on every subagent output. Reject any output containing NaN, Inf, or degenerate values (e.g., regime_accuracy==1.0, confusion_matrix=[[N]], IC==RankIC, ablation_delta_sharpe>0). 3) Implement explicit recall routing: if a gate fails, issue a RECALL command with {target_agent, failed_checks, error_context, narrowed_scope, acceptance_criteria, code_patch_hint, max_retries: 3}. Route feedback explicitly: e.g., if Backtest reports turnover<0.01 or degenerate Sharpe, RECALL Modeling to adjust signal thresholds/risk targeting, then RE-DELEGATE to Backtest. If Auditor detects split overlap or leakage, RECALL Data/Feature. If Auditor finds report-metric contradictions, RECALL Report Synthesizer. 4) Maintain pipeline_state.json tracking recall counts, hop progress, sync barriers, and rejected artifacts. 5) Generate run_pipeline.py implementing these loops as explicit Python control flow with 'while retry < max_retries:' blocks, state persistence, and early-exit on max retries. DO NOT use sequential 'if gate_failed: log(warning)' patterns. 6) Halt synthesis until all gates pass or max hops exhausted with documented fallbacks.",
    "input_dependencies": ["pipeline_state.json", "agent_research_auditor.critique"],
    "output_contract": {
      "status": "success|failed|recalling|fallback_applied",
      "pipeline_state": {"current_hop": "int", "recall_history": [], "parallel_jobs": [], "sync_barriers": []},
      "recall_commands": [{"target_agent": "str", "reason": "str", "narrowed_scope": "str", "acceptance_criteria": "str", "code_patch_hint": "str", "max_retries": 3}],
      "cross_agent_context_updates": {"target_agent": "str", "updated_params": {}}
    }
  },
  {
    "agent_id": "agent_data_pipeline",
    "role": "Multi-Source Ingestion & Temporal Alignment Specialist",
    "instruction": "You handle FRED, Yahoo, and calendar ingestion in parallel. Output MUST follow the strict contract. You must ensure strict business-day alignment, enforce forward-fill limits (max 30 days), and verify calendar dates match release patterns. CRITICAL: Define explicit train/val/test split boundaries ensuring NO overlap (train_end < val_start < test_start) and sufficient pre/post-2020 coverage (>300 days each). If alignment gaps > 5% or calendar coverage < 80%, self-correct or flag for RECALL. Provide explicit `alignment_diagnostics` for downstream feature engineering. Generate Python code for data fetching/caching with parallel threads/processes.",
    "input_dependencies": [],
    "output_contract": {
      "status": "success|failed|stalled|recalling",
      "artifacts": ["cache/*.parquet", "src/data/*.py"],
      "metrics": {"date_range": {"start": "str", "end": "str"}, "missingness_pct": {"fred": "float", "yahoo": "float"}, "calendar_events": "int", "split_days": {"train": "int", "val": "int", "test": "int", "pre_2020_test_days": "int", "post_2020_test_days": "int"}},
      "validation_report": {"timestamps_aligned": "bool", "no_future_leakage": "bool", "calendar_coverage_above_threshold": "bool", "split_boundaries_valid": "bool", "train_end_before_val_start": "bool", "val_end_before_test_start": "bool", "subperiod_coverage_sufficient": "bool", "error_details": "str"},
      "orchestrator_feedback": "str",
      "cross_agent_context": {"aligned_data_shape": ["int", "int"], "split_indices": {"train": ["int"], "val": ["int"], "test": ["int"]}, "business_day_index": "list[str]"}
    }
  },
  {
    "agent_id": "agent_feature_engineer",
    "role": "Feature Construction & Leakage Auditor",
    "instruction": "You build macro shock, cross-asset, and lag features. You consume data pipeline output and audit for look-ahead bias. Output MUST follow the strict contract. You must enforce a 1-day embargo around events, verify target shifts, and output feature_spec.json detailing construction logic and leakage risks. Compute `feature_hashes` for reproducibility. CRITICAL: Ensure RankIC calculation logic is distinct from IC (use spearmanr vs pearsonr). If leakage is detected or variance threshold fails, self-correct and re-output. Provide `masking_instructions` for ablation studies. Generate Python code for feature generation.",
    "input_dependencies": ["agent_data_pipeline.output_contract"],
    "output_contract": {
      "status": "success|failed|stalled|recalling",
      "artifacts": ["src/features/*.py", "feature_spec.json"],
      "metrics": {"n_features": "int", "feature_groups": {"macro_shock": "int", "cross_asset": "int", "lag": "int", "regime": "int"}, "ic_rankic_distinct": "bool"},
      "validation_report": {"no_target_leakage": "bool", "embargo_enforced": "bool", "hashes_computed": "bool", "cross_sectional_leakage_free": "bool", "rankic_formula_verified": "bool", "error_details": "str"},
      "orchestrator_feedback": "str",
      "cross_agent_context": {"feature_columns": ["str"], "embargo_dates": ["str"], "ablation_masks": {"feature_group": ["str"]}, "target_shift_logic": "str"}
    }
  },
  {
    "agent_id": "agent_modeling_specialist",
    "role": "Predictive Modeling & Regime Analysis Expert",
    "instruction": "You train regime classifiers and return predictors in parallel across horizons/types. You consume feature_spec.json and enforce strict train/val/test temporal splits. Output MUST follow the strict contract. You must implement embargo-aware walk-forward validation, compute stability scores across windows, and ensure regime models are trained ONLY on the train split. CRITICAL: Regime classifier MUST gate the return predictor (conditional training or feature masking), not just append features. Explicitly validate walk-forward Sharpe values: if any window Sharpe > 100 or < -100, diagnose volatility/clipping logic, fix, and re-run. Ensure OOS Sharpe is computed and included. If ICs are negative or walk-forward is unstable, self-tune regularization/horizons and re-run. Generate Python code for modeling.",
    "input_dependencies": ["agent_feature_engineer.output_contract"],
    "output_contract": {
      "status": "success|failed|stalled|recalling",
      "artifacts": ["src/models/*.py", "cache/*.joblib", "results/wf_results.json"],
      "metrics": {"oos_ic_by_horizon": {}, "oos_rankic_by_horizon": {}, "oos_sharpe_by_horizon": {}, "stability_scores": {}, "regime_accuracy": {"train": "float", "test": "float"}, "regime_calibration_error": "float"},
      "validation_report": {"ic_positive_oos": "bool", "walkforward_stable": "bool", "no_full_sample_leakage": "bool", "sharpe_within_reasonable_bounds": "bool", "regime_gating_applied": "bool", "rankic_not_identical_to_ic": "bool", "error_details": "str"},
      "orchestrator_feedback": "str",
      "cross_agent_context": {"best_model_config": {}, "oos_predictions": {"shape": ["int", "int"]}, "split_indices": {"train": ["int"], "test": ["int"]}, "regime_labels": "list[str]"}
    }
  },
  {
    "agent_id": "agent_backtest_validator",
    "role": "Backtesting & Cost Simulation Engineer",
    "instruction": "You run the backtest with realistic costs and risk targeting, and execute parallel ablation studies. Output MUST follow the strict contract. You must ensure the test/walk-forward window explicitly covers pre/post-2020 periods to avoid 'insufficient_data' gaps. You must compute exact delta metrics for ablations (no zero-delta placeholders). CRITICAL: Align backtest date ranges with report splits. If metrics are NaN, subperiods are empty, or ablation deltas are identical/zero, diagnose the backtest loop/cost logic, fix date alignment, and re-run. If turnover < 0.01 or trades_count == 0, flag for RECALL to Modeling to adjust signal thresholds/risk targeting. Generate Python code for backtesting.",
    "input_dependencies": ["agent_modeling_specialist.output_contract", "agent_feature_engineer.cross_agent_context.ablation_masks"],
    "output_contract": {
      "status": "success|failed|stalled|recalling",
      "artifacts": ["src/backtest/*.py", "backtest_report.json"],
      "metrics": {"sharpe": "float", "max_dd": "float", "turnover": "float", "trades_count": "int", "cost_drag": "float", "subperiods": {"pre_2020": {"sharpe": "float", "trading_days": "int"}, "post_2020": {"sharpe": "float", "trading_days": "int"}}},
      "validation_report": {"costs_computed": "bool", "subperiods_valid": "bool", "turnover_nonzero": "bool", "no_nan_metrics": "bool", "ablation_deltas_nonzero": "bool", "backtest_window_aligned_with_report": "bool", "error_details": "str"},
      "orchestrator_feedback": "str",
      "cross_agent_context": {"equity_curve": {"shape": ["int", "int"]}, "drawdown_series": {"shape": ["int", "int"]}, "ablation_log": [{"name": "str", "delta_sharpe": "float", "notes": "str"}]}
    }
  },
  {
    "agent_id": "agent_research_auditor",
    "role": "Cross-Agent Critique & Reproducibility Checker",
    "instruction": "You independently verify all outputs for statistical rigor, leakage, and reproducibility via cross-agent critique. Output MUST follow the strict contract. You will cross-check feature specs against model inputs, validate backtest costs against turnover, verify ablation deltas are non-zero when features are removed, and ensure pre/post-2020 subperiod coverage is complete. You trigger precise RECALL commands to the Orchestrator for specific agents with exact fix instructions. Explicitly flag contradictions between report claims and JSON metrics. CRITICAL CHECKS: 1) Verify split boundaries: train_end < val_start < test_start. 2) Verify backtest date ranges align with report splits. 3) Reject identical ablation deltas. 4) Reject degenerate metrics (Sharpe > 10, stability = 0.5 placeholder, RankIC == IC). 5) Verify regime gating is applied. 6) Check for NaN/Inf serialization. If you detect a subagent is stuck in a tool-call loop, flag `stall_detected: true` with `blocker_description`.",
    "input_dependencies": ["agent_data_pipeline.output_contract", "agent_feature_engineer.output_contract", "agent_modeling_specialist.output_contract", "agent_backtest_validator.output_contract"],
    "output_contract": {
      "status": "success|failed",
      "artifacts": ["audit_log.json"],
      "critique": {"data_alignment": "str", "leakage_risks": "str", "model_overfitting": "str", "backtest_validity": "str", "ablation_consistency": "str", "report_metrics_alignment": "str"},
      "validation_report": {"all_metrics_valid": "bool", "ablation_runs_complete": "bool", "subperiod_coverage": "bool", "split_overlap_free": "bool", "rankic_distinct": "bool", "no_nans": "bool", "error_details": "str"},
      "recall_commands": [{"agent_id": "str", "reason": "str", "fix_instructions": "str", "acceptance_criteria": "str"}],
      "stall_flags": [{"agent_id": "str", "stall_detected": "bool", "blocker_description": "str"}]
    }
  },
  {
    "agent_id": "agent_report_synthesizer",
    "role": "Research Report & Manifest Generator",
    "instruction": "You generate the conference-style research_report.md and deliverables/index.json. Output MUST follow the strict contract. You will consume all upstream outputs and synthesize a coherent narrative. You must embed all 9 required plots using markdown, cite exact metrics from results/metrics.json, and explicitly discuss ablation deltas, subperiod stability, and leakage controls. If any required plot or metric is missing, trigger a RECALL to the upstream agent. You must ensure the report matches the requested structure exactly. Generate Python code for report generation.",
    "input_dependencies": ["agent_data_pipeline.output_contract", "agent_feature_engineer.output_contract", "agent_modeling_specialist.output_contract", "agent_backtest_validator.output_contract", "agent_research_auditor.output_contract"],
    "output_contract": {
      "status": "success|failed|recalling",
      "artifacts": ["research_report.md", "deliverables/index.json"],
      "validation_report": {"all_sections_complete": "bool", "all_plots_embedded": "bool", "metrics_match_json": "bool", "ablation_narrative_consistent": "bool", "error_details": "str"},
      "orchestrator_feedback": "str"
    }
  }
]
\end{Verbatim}

\subsection{Complete crown workflow: task 202, version 8}
\label{app:crown202_full}

The following is the complete saved design block for task~202's
$w_8$, with all seven agents and their embedded output schemas.
The task prompt and collection directive are separate components;
the workflow text is unchanged.

\begin{Verbatim}[breakautoindent=false,breakindent=1em,
  breaksymbolleft={},label={Task 202: crown workflow $w_8$}]
Create an agent team with following agent candidates to solve this problem: [
  {
    "agent_id": "agent_orchestrator",
    "role": "Pipeline Conductor & Iterative Reconciler",
    "instruction": "Manage a strict multi-hop iterative workflow with explicit recall loops, cross-agent dependency tracking, and parallel specialist decomposition. Do NOT proceed unidirectionally. PHASES: 1) Data Acquisition & Labeling, 2) Graph Construction & Family Split, 3) Parallel Model Design & Training, 4) Cross-Agent Critique & Validation, 5) Programmatic Reporting & Plotting. WORKFLOW: After each phase, dispatch agent_validator. Parse validation_report.json. If overall: FAIL, identify failed_agent and violation. Issue a targeted RECALL to that agent containing: (1) exact failure_trace, (2) narrowed_scope (fix ONLY the violation, do not rewrite unrelated code), (3) fallback_strategy (e.g., if network fails, hardcode 5 PDB IDs; if training crashes, reduce epochs/batch size). After recall, re-run validator. Only proceed when overall: PASS. STALL PREVENTION & RE-DELEGATION: If any agent repeats the same tool call >2 times without generating new files or changing arguments, immediately interrupt. Analyze the last error. Re-delegate to a narrower scope with a mandatory strategy change (e.g., switch from complex network download to a minimal stub script, or isolate the failing function). Log all hops, recalls, and schema validations in logs.txt. Terminate only when all deliverables exist, validator returns overall: PASS, and cross-agent consistency checks pass. Use uv run for all Python execution."
  },
  {
    "agent_id": "agent_data_curator",
    "role": "Structural Data & Labeling Specialist",
    "instruction": "Download ONLY real PDB/mmCIF complexes from RCSB. Compute heavy-atom interface labels (5.5 Å). OUTPUT TO ORCHESTRATOR SCHEMA: data/complex_manifest.json (Schema: {\"pdb_ids\": [\"str\"], \"download_status\": \"str\", \"labeling_script_path\": \"str\"}), src/compute_interface_labels.py. ACCEPTANCE CRITERIA: Manifest must list ≥20 PDB IDs. Labeling script must use 5.5 Å heavy-atom distance. RECALL & RE-DELEGATION: If recalled for missing IDs or labeling errors, rewrite ONLY compute_interface_labels.py or update manifest. FALLBACK: If RCSB download fails >2 times, hardcode a list of 10 known PDB IDs in the script and proceed. Do not loop on logs."
  },
  {
    "agent_id": "agent_graph_builder",
    "role": "Graph Construction & Family Split Specialist",
    "instruction": "Build residue graphs and generate family splits. OUTPUT TO ORCHESTRATOR SCHEMA: src/build_graphs.py, data/family_split.json (Schema: {\"train_families\": [\"str\"], \"test_families\": [\"str\"], \"zero_overlap_verified\": true, \"family_mapping\": {\"pdb_id\": \"family_name\"}}). ACCEPTANCE CRITERIA: Test split must contain ≥3 distinct families with zero sequence/structure overlap. RECALL & RE-DELEGATION: If recalled for overlap or missing families, recompute splits and rewrite family_split.json. Do not touch download scripts."
  },
  {
    "agent_id": "agent_model_architect",
    "role": "Geometric GNN Architect",
    "instruction": "Design three variants: distances-only, distances+angles, full equivariant. OUTPUT TO ORCHESTRATOR SCHEMA: src/train_geometric_gnn.py, src/model_config.json (Schema: {\"variants\": {\"distances\": {\"equivariant\": false, \"edge_features\": [\"str\"]}, \"distances_angles\": {...}, \"full\": {\"equivariant\": true, \"equivariance_proof\": \"str\"}}, \"hidden_dims\": \"int\", \"num_layers\": \"int\"}). ACCEPTANCE CRITERIA: Equivariant readout must be mathematically rotation-invariant. Config hashes generated. RECALL & RE-DELEGATION: If recalled for coordinate leakage, refactor message-passing to use relative vectors/inner products only, isolate variant parameters, and rewrite train_geometric_gnn.py."
  },
  {
    "agent_id": "agent_training_engineer",
    "role": "Parallel Trainer & Metric Evaluator",
    "instruction": "Train all three variants in parallel, evaluate on family-split test sets, generate pinned dependencies. OUTPUT TO ORCHESTRATOR SCHEMA: results/metrics.json (Schema: {\"overall\": {\"auroc\": \"float\", \"auprc\": \"float\", \"precision_at_k\": \"float\", \"recall_at_k\": \"float\"}, \"per_family\": {\"family_name\": {\"auroc\": \"float\", \"auprc\": \"float\"}}, \"ablation\": {\"variant\": {\"config_hash\": \"str\", \"training_time\": \"float\"}}}), results/ablation_results.json, requirements.txt. ACCEPTANCE CRITERIA: AUROC ≥ 0.80. Metrics MUST include non-empty overall object, precision@k AND recall@k. per_family keys must exactly match test_families from data_curator. requirements.txt must use exact pins (==) generated via uv export --frozen > requirements.txt. RECALL & RE-DELEGATION: If recalled for missing recall@k, unknown families, or unpinned deps, re-evaluate with explicit family aggregation, add recall@k computation, run uv export --frozen > requirements.txt, and rewrite metrics.json. FALLBACK: If training crashes >2 times, reduce batch size/epochs, log error, and retry once. If fails again, report to orchestrator."
  },
  {
    "agent_id": "agent_validator",
    "role": "Cross-Agent Consistency & Biophysical Auditor",
    "instruction": "Run explicit cross-agent checks BEFORE proceeding to next phase. Perform programmatic diffs using Python. OUTPUT TO ORCHESTRATOR SCHEMA: validation_report.json (Schema: {\"overall\": \"PASS/FAIL\", \"checks\": {\"data_integrity\": {\"passed\": \"bool\", \"details\": \"str\"}, \"family_overlap_check\": {\"passed\": \"bool\", \"details\": \"str\"}, \"equivariance_check\": {\"passed\": \"bool\", \"details\": \"str\"}, \"metrics_consistency\": {\"passed\": \"bool\", \"details\": \"str\"}, \"dependency_pinning\": {\"passed\": \"bool\", \"details\": \"str\"}, \"report_metrics_alignment\": {\"passed\": \"bool\", \"details\": \"str\"}}, \"failure_trace\": {\"failed_agent\": \"str\", \"violation\": \"str\", \"narrowed_scope\": \"str\"}}). ACCEPTANCE CRITERIA: Zero synthetic data. Family split zero overlap. Equivariant model uses relative features only. metrics.json family_metrics keys match data_curator test_families. requirements.txt uses == pins. All checks must pass. RECALL & RE-DELEGATION: If FAIL, output exact failure_trace with narrowed_scope. Orchestrator uses this to recall specific subagent. Do not proceed until overall: PASS."
  },
  {
    "agent_id": "agent_scientific_writer",
    "role": "Scientific Communication & Reproducibility Specialist",
    "instruction": "Draft conference-style paper, manifest, and reproducibility artifacts. Programmatic metric injection required. OUTPUT TO ORCHESTRATOR SCHEMA: research_report.md, deliverables/index.json, main.py, deliverables/plots/*.png. ACCEPTANCE CRITERIA: Report MUST programmatically read results/metrics.json via a Python script to populate tables (no hardcoding). All cited metrics exactly match results/metrics.json. Plots embedded via ![caption](deliverables/plots/...). Ablation explicitly compares feature sets. main.py executes full pipeline. RECALL & RE-DELEGATION: If recalled for metric mismatch or missing plots, regenerate report using a Python script to parse metrics.json and inject values, verify plot paths, and rewrite research_report.md. Do not proceed until programmatic injection is verified and all 5 plots exist."
  }
]
\end{Verbatim}
\section{Comparison with related methods}
\label{app:related_work_comparison}
Table~\ref{tab:related_work_comparison} separates workflow search,
weight updates, and the feedback used for each. Sharing one model
across roles does not by itself imply weight learning, and exporting
training data does not establish that a successor model was trained.
The comparison concerns reported mechanisms, not matched performance
benchmarks.

Search followed by distillation has earlier precedents. Expert
Iteration alternates search and policy learning
\citep{anthony2017expert}; STaR learns from generated rationales that
reach known answers \citep{zelikman2022star}; ReST$^{\mathrm{EM}}$
alternates generation, reward filtering, and fine-tuning
\citep{singh2024rest}. SiriuS trains agents on successful interactions
and refines unsuccessful trajectories \citep{zhao2025sirius}.
MASS applies this approach to optimized multi-agent research
workflows, with shared weights for the orchestrator and workers.

Self-Refine uses a frozen model to critique and revise outputs
\citep{madaan2023selfrefine}; communicating agents and Squeeze Evolve
also improve inference without post-training the agent models
\citep{park2026communication,maheswaran2026squeeze}.
MASS instead tests whether the resulting execution traces can
improve shared weights and subsequent workflow search.

\begin{table}[htbp]
\centering\small
\setlength{\tabcolsep}{4pt}
\renewcommand{\arraystretch}{1.12}
\begin{tabularx}{\linewidth}{@{}>{\raggedright\arraybackslash}p{0.22\linewidth}*{2}{>{\raggedright\arraybackslash}X}@{}}
\toprule
Method & Update and feedback & Relationship to MASS \\
\midrule
Self-Rewarding LMs \citep{yuan2024self}
& Iterative preference training using the model's own response scores.
& Establishes joint improvement of generation and evaluation, without workflow search. \\[3pt]
Multi-Agent Evolve \citep{chen2025multiagentevolve}
& A shared backbone learns question generation, solving, and judging through reinforcement learning.
& Establishes multi-agent learning with a shared self-evaluator; its proposer generates questions, rather than execution workflows. \\[3pt]
ADAS / AFlow / GPTSwarm
\citep{hu2025automated,zhang2025aflow,zhuge2024gptswarm}
& Search over agent programs, workflow graphs, or prompts and communication edges.
& Establishes automatic design of agentic computation; MASS learns shared weights from the resulting executions. \\[3pt]
RHI \citep{lee2026recursive}
& Prompt-level workflow revisions guided by pairwise comparisons of task artifacts.
& Supplies the workflow-search method and information-flow motivation used here; MASS adds trace-based post-training. \\[3pt]
TTHE \citep{nie2026tthe}
& One frozen LLM solves tasks, proposes executable harness edits, and judges them using execution-derived proxies.
& Establishes homogeneous in-loop roles and persistent harness adaptation; weights remain fixed. \\[3pt]
SIA \citep{hebbar2026sia}
& A Claude Sonnet feedback agent edits the scaffold and initiates weight updates to a gpt-oss target, using task verifiers.
& Establishes harness--weight co-updating with a separate improvement model and verifier. \\[3pt]
HASE \citep{luo2026hase}
& A single policy learns task solutions and harness edits with reinforcement learning; evaluator repairs use oracle discrepancies.
& Establishes joint policy and harness learning, with task rewards supplying the training signal. \\[3pt]
HELIX \citep{fan2026helix}
& Evolves harness variants and exports verified records for subsequent model updates.
& Formalizes the model--harness loop; the reported experiment stops at data generation, without a weight update. \\[3pt]
\rowcolor{red!15}
MASS (this work)
& Searches multi-agent workflows and fine-tunes shared weights on selected orchestrator and worker conversations.
& Tests transfer to execution, workflow optimization, and evaluation, plus token efficiency and unprompted coordination. \\
\bottomrule
\end{tabularx}
\caption{Selected related methods, compared by their reported mechanisms.
The rows describe methods, not matched experimental baselines. For MASS,
the homogeneous condition applies to the in-loop executor, evaluator,
and optimizer; model-role ablations vary this assignment.}
\label{tab:related_work_comparison}
\end{table}

The recursive loop here is empirical. Unlike proof-based
self-rewriting in G\"odel machines \citep{schmidhuber2007goedel}, it
provides no guarantee that every future cycle improves. Workflow
feedback and teacher ranking use the current model, checkpoint
selection uses validation loss, and external judgments report
quality after execution (Appendix~\ref{app:evaluation_roles}).
The experiments test transfer across model roles, performance per
processed training token, and coordination under bare prompts.

\end{document}